\documentclass[journal]{IEEEtran}
\IEEEoverridecommandlockouts
\usepackage{graphicx}
\usepackage{amssymb,amsfonts}
\usepackage{booktabs}
\usepackage{multirow}
\usepackage{epstopdf}
\usepackage{textcomp}
\usepackage{times}
\usepackage{amsmath}
\usepackage{algorithm}
\usepackage{algorithmicx}
\usepackage{algpseudocode}
\usepackage{subfigure}
\usepackage{bm}
\usepackage{float}
\usepackage{cases}
\usepackage{enumerate}
\usepackage[numbers]{natbib}
\usepackage{multicol}
\usepackage{paralist}
\usepackage{mathrsfs}

\begin{document}

\title{SBPF: Sensitiveness Based Pruning Framework For Convolutional Neural Network On Image Classification}
\author{Yiheng~Lu,
	Maoguo~Gong, \IEEEmembership{Senior Member,~IEEE},
	Wei~Zhao,
	Kai-yuan~Feng, \IEEEmembership{Student Member, IEEE}
  and Hao~Li, \IEEEmembership{Member, IEEE}

\thanks{This work was supported in part by the National Natural Science Foundation
of China under Grant 62036006, Grant 61906146, Grant 61876144 and Grant 61876145. \emph{(Corresponding author: Maoguo Gong.)}}
\thanks{Yiheng Lu and Wei Zhao are with the State Key Laboratory of Integrated Services Networks, School of Computer Science and Technology, Xidian University, Xi'an 710071, China (e-mail: lyhxdu@gmail.com; ywzhao@mail.xidian.edu.cn)}
\thanks{Maoguo Gong, Kaiyuan Feng and Hao Li are with the Key Laboratory of Intelligent Perception
and Image Understanding, School of Electronic Engineering, Ministry of
Education, Xidian University, Xi'an 710071, China (e-mail: gong@ieee.org; fkylwl@gmail.com; omegalihao@gmail.com)}
}

	\maketitle
\begin{abstract}
\label{abstract}
Pruning techniques are used comprehensively to compress convolutional neural networks (CNNs) on image classification.
However, the majority of pruning methods require a well pre-trained model to provide useful supporting parameters, such as $\mathscr{C}_1$-norm, BatchNorm value and gradient information, which may lead to inconsistency of filter evaluation if the parameters of the pre-trained model are not well optimized. Therefore, we propose a sensitiveness based method to evaluate the importance of each layer from the perspective of inference accuracy by adding extra damage for the original model. Because the performance of the accuracy is determined by the distribution of parameters across all layers rather than individual parameter, the sensitiveness based method will be robust to update of parameters. Namely, we can obtain similar importance evaluation of each convolutional layer between the imperfect-trained and fully trained models. For VGG-16 on CIFAR-10, even when the original model is only trained with 50 epochs, we can get same evaluation of layer importance as the results when the model is trained fully. Then we will remove filters proportional from each layer by the quantified sensitiveness. Our sensitiveness based pruning framework is verified efficiently on VGG-16, a customized Conv-4 and ResNet-18 with CIFAR-10, MNIST and CIFAR-100, respectively.

\end{abstract}


%



\section{Introduction}
\label{introduction}
\IEEEPARstart{C}{onvolutional} neural networks have been widely adopted in the field of computer vision, e.g., image classification \cite{kriz:ima,googlenet,googlenetv4,densnet}, object detection \cite{girshick-et-al:richfeature,fasterrcnn,maskrcnn} and semantic segmentation \cite{jonathan-et-al:fcnss}.
The increase of width and depth for CNNs is proved effectively to achieve state-of-the-art performance on various tasks \cite{hekaiming:resnet,karen:vdc}. Recently, He \emph{et al.} \cite{hekaiming:resnet} proposed ResNets with residual units to solve degradation problem caused by VGGNets \cite{karen:vdc}, which led to an impressive result with 152 convolutional layers. However, these networks with multi-convolutional layers are verified to be too deep to train efficiently. Namely, complicated structure can reinforce representation power on huge datasets, but inevitably result in high memory footprint and computation consumption during training and inference process. For example, the FLOPs of VGG-16 are up to $0.31$ Giga when it processes a $32\times32$ image with three channels. Consequently, pruning techniques are proposed to improve efficiency by removing redundant parameters or filters.

 Originally, unstructured pruning methods have been explored broadly in  \cite{Yann:obd,babak:obu,sparsity:soravit}, which removed partial parameters by evaluating the importance of corresponding neurons or connections.
However, unstructured pruning cannot reduce FLOPs consumption apparently because the number of feature maps for each convolutional layer cannot be eliminated. Therefore, structured pruning methods were proposed to remove intact filters at layer-level, which could reduce FLOPs apparently. For example, heuristic methods such as \cite{nn-filter-grouping,haoli:pffec,hengyuan:datadriven} and automatic methods such as \cite{pavlo:nvidia,zhuangliu:networkslimming,zehao:datadriven} evaluated the importance of each filter through corresponding weight value. They then slimmed networks by removing trivial filters, and consequently, could yield better performance than unstructured pruning methods.
 \begin{table*}[htbp]
\tiny
  \centering
  \caption{The pruning for VGG-16 on CIFAR-10 with $65\%$ pruning ratio. Accu-O and Accu-P stands for the accuracy of original and pruned model, respectively. Both Taylor-Pruning and BN-pruning shows inconsistency when VGG-16 are trained differently.}
  \setlength{\tabcolsep}{1.6mm}{
        \begin{tabular}{|p{1.7cm}|p{1.2cm}|p{1.2cm}|p{0.8cm}|p{0.5cm}|p{0.5cm}|p{0.5cm}|p{0.5cm}|p{0.5cm}|p{0.5cm}|p{0.5cm}|p{0.5cm}|p{0.5cm}|p{0.5cm}|p{0.5cm}|p{0.5cm}|p{0.5cm}|p{0.8cm}|}
    \hline
    Models/Pruning Ratio & Training Epochs & Methods &\textbf{Accu-O} & \multicolumn{13}{c}{Layers}&\textbf{Accu-P} \\
    \hline
          &       &       &       & Conv1 & Conv2 & Conv3 & Conv4 & Conv5 & Conv6 & Conv7 & Conv8 & Conv9 & Conv10 & Conv11 & Conv12 & Conv13 & \\
    \cline{2-18}

    \multirow{6}[0]{*}{VGG-16 (0.65)} & 50    & \multirow{3}[0]{*}{Taylor-Pruning} & 0.902 & 36    & 43    & 62    & 67    & 83    & 93    & 110   & 105   & 141   & 197   & 152   & 202   & 188 &0.891 \\
    \cline{2-2}\cline{4-18}
          & 100   &       & 0.92  & 42    & 42    & 70    & 66    & 91    & 82    & 105   & 104   & 137   & 199   & 171   & 183   & 187 &0.900\\
          \cline{2-2}\cline{4-18}
          & 150   &       & 0.92  & 43    & 41    & 61    & 54    & 81    & 84    & 115   & 119   & 163   & 198   & 169   & 198   & 153 &0.905\\
          \cline{2-18}
          & 50    & \multirow{3}[0]{*}{BN-Pruning} & 0.902 & 58    & 60    & 123   & 123   & 250   & 250   & 250   & 196   & 26    & 14     & 11     & 14     & 103&0.911 \\
          \cline{2-2}\cline{4-18}
          & 100   &       & 0.92  & 54    & 62    & 125   & 125   & 256   & 250   & 250   & 215   & 28    & 7     & 6     & 9     & 91& 0.915\\
          \cline{2-2}\cline{4-18}
          & 150   &       & 0.92  & 52    & 64    & 128   & 128   & 256   & 256   & 250   & 205   & 38    & 9     & 2     & 4     & 85 & 0.920\\
          \hline
    \end{tabular}}%
  \label{pruning_specific}%
\end{table*}%

 However, one drawback that existed in structured pruning methods should be concerned: the importance evaluation of filters are implemented only depending on weight value of original models, which may twist the pruning result when the parameters of the original model are not distributed well, namely, the filters pruning varies with the update of parameters for individual filter.

 To reveal this issue, we compare two prevalent weight value based structured pruning methods \cite{pavlo:nvidia,zhuangliu:networkslimming} on the same network and output the pruning results when the original model is trained differently. Specifically, the "Taylor-Pruning" in \cite{pavlo:nvidia} decomposed network pruning process into polynomials with first order, which simplified filter importance evaluation as the product of value and gradient of the output features. The "BN-Pruning" in \cite{zhuangliu:networkslimming} applied a sparsity coefficient on BatchNorm layer then constrained trivial neurons to be zero, which would instruct the filters in corresponding convolutional layer to be pruned. The results are displayed on Table \ref{pruning_specific}. Clearly, the number of filters within each layer were removed differently when the original model was trained with different epochs even under same pruning ratio, then the recovered accuracy of the pruned model will be damaged obviously. Especially, even the accuracy of original models reach to $92\%$ both at 100 and 150 training epochs, the filters are still pruned variously, which damage the representation ability of the pruned model. The main reason for above dilemma can be explained that the weight-oriented methods are sensitive to the update of parameters, then lead to the misjudgement of filter importance. Therefore, we put focus on evaluating the importance of each layer rather than individual filters.

In this paper, we propose a sensitiveness based method to describe the importance of each layer from the perspective of reliability and stability, which can be measured by the accuracy gap between original and parameters-sabotaged and structure-sabotaged models, respectively. Specifically, the parameters-sabotaged model can be obtained by freezing the parameters within one specified layer and keep other layers unchanged, then the parameters-sabotaged model will be retrained with one epoch to observe the variation of inference accuracy. The structure-sabotaged model is yielded by removing filters randomly from one layer while maintaining other layers, and the structure-sabotaged model will be retrained fully to calculate the change of inference accuracy compared with original model. Notably, both the reliability and stability are measured by imposing the resistance on the distribution of parameters, which lead to the change of the inference accuracy. Therefore, reliability and stability are robust to the update of parameters for individual filter. Simultaneously, we adopt the vMF distribution in \cite{visualizing} to illustrate that the evaluation of sensitiveness for each layer can keep consistent between the well-trained and the imperfect-trained models. Especially, the imperfect-trained models are defined as the models that are trained after a quarter of total epochs. Usually, the parameters of a model are updated rapidly at the beginning of the training stage, then the variation of the parameters distribution across all layers will be smooth during the following training process, especially for adjacent training epochs.  Then the difference within all layers will be greater than the discrepancy between the well-trained and the imperfect-trained models, namely, we could obtain similar evaluation of sensitiveness for each layer when the original model is trained with different epochs.

 After evaluation, the layers with similar sensitiveness will be clustered as a group, and same pruning ratio will be assigned to the layers within same group. In order to obtain better pruning results, the allocation of pruning ratio will be divided into several iterations rather once time, and the slimming process will be ended until the overall pruning ratio reaches to the pre-defined ratio. The efficiency of the proposed framework is verified on VGGNets and ResNet-18 with different datasets. For example, the FLOPs can be reduced almost 40\% on VGG-16 and 50\% on ResNet-18 with CIFAR-10 and CIFAR-100, respectively. Besides, 85.65\% parameters are reduced for VGG-16 on CIFAR-10. There are two main contributions in this paper:
\begin{compactenum}
\item We observe the issue of inconsistency for filters importance evaluation that appears in previous weight value based pruning methods, and rethink the dominated status of network structure.

\item To the best knowledge of us, we are the first to propose a sensitiveness based pruning framework to evaluate the importance of each convolutional layer, which is robust to the update of weight value for individual filter. Then filters will be removed according to the sensitiveness. Consequently, identical slimmed models can be yielded even when the original models are trained with different epochs.

\end{compactenum}
This paper is organized as follows: Section II describes the background and related work about pruning methods for neural networks. Section III introduces the details about our sensitiveness based pruning framework. Section IV gives the results of different experiments. In Section V, the comparison with previous methods and details are analyzed, and the introduced hyper-parameter is also discussed. Finally, the conclusion is presented in Section VI.

\section{Related work}
\subsection{Pruning Techniques}
Pruning techniques could reduce the size of hand-craft networks from top-to-bottom. Moreover, networks pruning can be classified into two main categories: unstructured pruning and structured pruning. The unstructured pruning methods focus on importance of partial parameters within each individual neuron. The original work of unstructured pruning was done by \cite{Yann:obd}, which pruned neurons under salience measure. Hassibi \emph{et al.} \cite{babak:obu} removed unimportant weights through second-order derivative information of neurons. And Srinivas \emph{et al.} \cite{datafree:suraj} applied data-free method to remove useless neurons. However, pruning only for fully connection layers ignored the influence of convolutional layers and could not speed up classical CNNs. Recently, Han \emph{et al.} \cite{song:cdnnptqhc} achieved great compression ratio on AlexNet \cite{kriz:ima} and VGGNets \cite{karen:vdc} through small magnitude pruning process. Louizos \emph{et al.} \cite{lsnn} used stochastic gate to learn sparse networks through $L_0$-norm regularization. Molchanov \emph{et al.} \cite{dropoutsaprse} applied dropout techniques both in fully connection and convolutional layers to reduce redundant connections. Wu \emph{et al.}
\cite{decomposition} reduced the computation costs by matrix factorization, which could promote sparse parameters highly in many popular networks. Another unstructured pruning job that was done by \cite{lottery} has attracted great attention by using winning tickets to initialize a pruned model, which would make training particularly effective. However, these non-structured methods can not reduce the number of feature maps across the whole network, which contribute less reduction for FLOPs consumption, as well as requiring special hardware to support the computation \cite{song:eie}.

Structured pruning methods could remove intact filters, which change the width of each convolutional layers while maintaining the depth of original network. Some heuristic methods were introduced to remove redundant filters by $\mathscr{C}_1$-norm \cite{haoli:pffec} and APoZ \cite{hengyuan:datadriven}, respectively. Besides, Yu \emph{et al.} \cite{NISP} and Luo \emph{et al.} \cite{Thinet} pruned filters in current layers with the assistance of next layers through weight information. Meanwhile, sparsity constraint has been applied on channels as scaling factor during training process, then the corresponding magnitude was used to instruct channel pruning \cite{zhuangliu:networkslimming}. Similarly, Zhou \emph{et al.} \cite{hanzhou:tcn} proposed group sparsity to smooth the pruning process after training. Wu \emph{et al.} \cite{nn-decom} investigated the effectiveness of hierarchical Tucker (HT) and tensor-train (TT) decomposition methods on deep neural networks, and the strategy with combined HT and TT has been verified more efficiently than one method only on convolutional neural network. In \cite{nn-structured}, two sets of binary random variables were introduced as switch-gates to choose useful hidden neurons from RNNs, and consequently, the pruning technique was proved possibly to be applied on non-CNNs or non-DNNs. Besides, global automatic pruning methods also performed efficiently on many task. Molchanov \emph{et al.} \cite{pavlo:nvidia} proposed a Taylor expansion pruning methods to compress models through several pruning iterations, and BatchNorm neurons could be scaled to zeros to instruct filters pruning globally \cite{zhuangliu:networkslimming,zehao:datadriven}. Another useful method using AutoML technique was meta-pruning \cite{metalearning}, which proposed an evolutionary procedure to search for good-performing pruned networks, and only little human interference were required. Besides, Aimar \emph{et al.} \cite{NullHop} proposed an efficient CNNs accelerator that was called NullHop to exploit the sparsity of neuron activation, which could be used to accelerate computation and reduce memory costs. Similarly, Shah \emph{et al.} \cite{coprocessor} proposed a field-programmable gate array with runtime programmable co-processor to speed up feed-forward computation of DNNs on FPGAs. All of these structured pruning methods achieved great acceleration of networks both on software and hardware, but were also weight value based methods, which may lead to the inconsistency of evaluation for filter importance.
\subsection{Prevalent compression techniques from top-to-bottom}
Matrix approximation, quantization and knowledge distillation are another three prevalent top-to-bottom network compression strategies for hand-craft networks. For matrix approximation reduction, Denton \emph{et al.} \cite{linear:denton} used low-rank approximation to reduce computation costs for linear layers. Cintra \emph{et al.} \cite{approximation} presented a post-training approximation scheme that could replace parameters of original network with low-complexity. Similarly, Quantization techniques were used to avoid float performing. For instance, HashNet \cite{quantization:weilin} and binarization \cite{binary:matthiew} were employed to quantify the weight of network. Besides, knowledge distillation could distill useful information from original teacher model to relative small student model. Hinton \emph{et al.} \cite{knowledgedistation} achieved surprising results by distilling knowledge in an ensemble of models into a single model. And Korattikara \emph{et al.} \cite{knowledgedistationbaysion} distilled a Monte Carlo approximation to the posterior predictive density into a more compact form, which required less computation, as well as maintaining preferable test accuracy. However, all these methods only focus on parameter of models while ignoring effect of structure of network, which may impede us to recognize the intrinsic importance of each convolutional layer.
\subsection{Customized model from bottom-to-top}
Bottom-to-top schedule may customize a network from some basic elements, e.g., filters, pool, and activation function. Correspondingly, computation occupation could be reduced at beginning. For example, original models were designed more compact to adapt mobile devices on various tasks\cite{MobileNetV1,MobileNetV2,MobileNetV3}. More recently, neural network architecture search (NAS) techniques \cite{NASreinforce,NASlearningtransferingable} were proposed to devise appropriate architecture for specified assignments, namely, an automatic customized network was more suitable for generalization than human designed model. But NAS techniques usually require massive GPUs-supported hardware to reach ultimate convergence, which is uneconomical and time costing.

Consequently, we claimed a  sensitiveness based pruning framework with multi-steps to obtain the importance of each convolutional layer without depending of weight information, which could maintain the consistency of evaluation for filter importance. Then the explored sensitiveness could be used to instruct overall pruning for over-parameterized models.
\begin{figure*}[!htbp]
  \centering
  \includegraphics[width=7.0in]{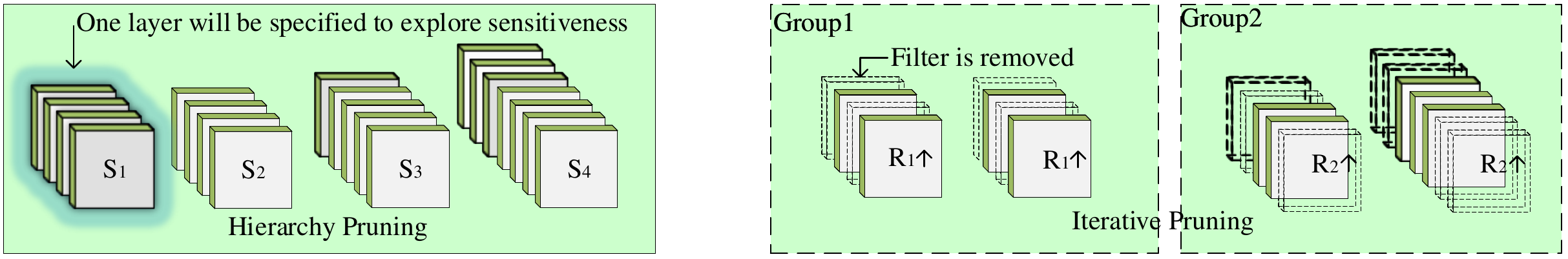}
  \caption{Multi-steps pruning procedures. $\emph{S}_l$ stands for the  sensitiveness of $l$-th convolutional layer. $\emph{R}_k$ stands for assigned pruning ratio for the $k$-th grouped layers. The layers within same group will be allocated with same pruning ratio, and the upper arrow means the pruning ratio will increase iteratively.}
  \label{overall_pruning_process}
\end{figure*}
\section{Multi-steps Network Pruning}
\label{methodology}
Show as Fig. \ref{overall_pruning_process}., the overall pruning process will be implemented through multi-steps with evaluation and slimming. At the left part in Fig. \ref{overall_pruning_process}., hierarchy pruning will be adopted to evaluate the sensitiveness of each layer through the reliability and stability that are obtained by parameter-sabotaged and structure-sabotaged models, respectively. Hierarchy pruning means the sensitiveness of each layer can be evaluated in parallel because every layer is explored as an individual item. In the right part of Fig. \ref{overall_pruning_process}., layers will be clustered into different groups according to the sensitiveness, and layers within one group will be allocated with same pruning ratio because they share similar sensitiveness. Then the original model will be pruned by the specified pruning ratio, as well as being retrained fully to observe whether the inference accuracy drops. If the inference accuracy is greater or equal to the baseline, each group will be reallocated different pruning ratio, namely, iterative pruning strategy will be applied to search for an appropriate slimmed model. Moreover, the consistency of reliability and stability will be analyzed by the vMF distribution when the original model is trained with different epochs.
 \subsection{Sensitiveness exploring during hierarchy pruning}
\subsubsection{Definition of Sensitiveness}
The sensitiveness of each layer can be defined as the sum value of reliability and stability, which will be explored by the recovering of accuracy when the parameter or structure of a specified layer is damaged while keeping other layers unchanged.

Reliability measures the variation of inference accuracy when the parameter distribution for a specified layer is interfered but with integral structure, which can be calculated by the accuracy gap between original and parameter-sabotaged model. The higher reliability means the specified layer will be more important. Specifically, we freeze the parameters of the specified layer while updating the parameters-sabotaged model with one retraining epoch, then the inference accuracy may be effected due to the change of parameters distribution. Stability measures the power of the structure-sabotaged model to recover the inference accuracy when the structure of a specified layer is damaged. Namely, filters will be removed randomly from a specified layer and other layers will keep unchanged. Then the structure-sabotaged model will be retrained with full training epochs to approach original capability. Clearly, if the structure-sabotaged model could restore baseline inference accuracy, the specified layer will be authorized with lower importance, and vice versa. Consequently, we will calculate stability by the difference between original and structure-sabotaged model.

\begin{figure}[!htbp]
  \centering
  \includegraphics[width=2.0in]{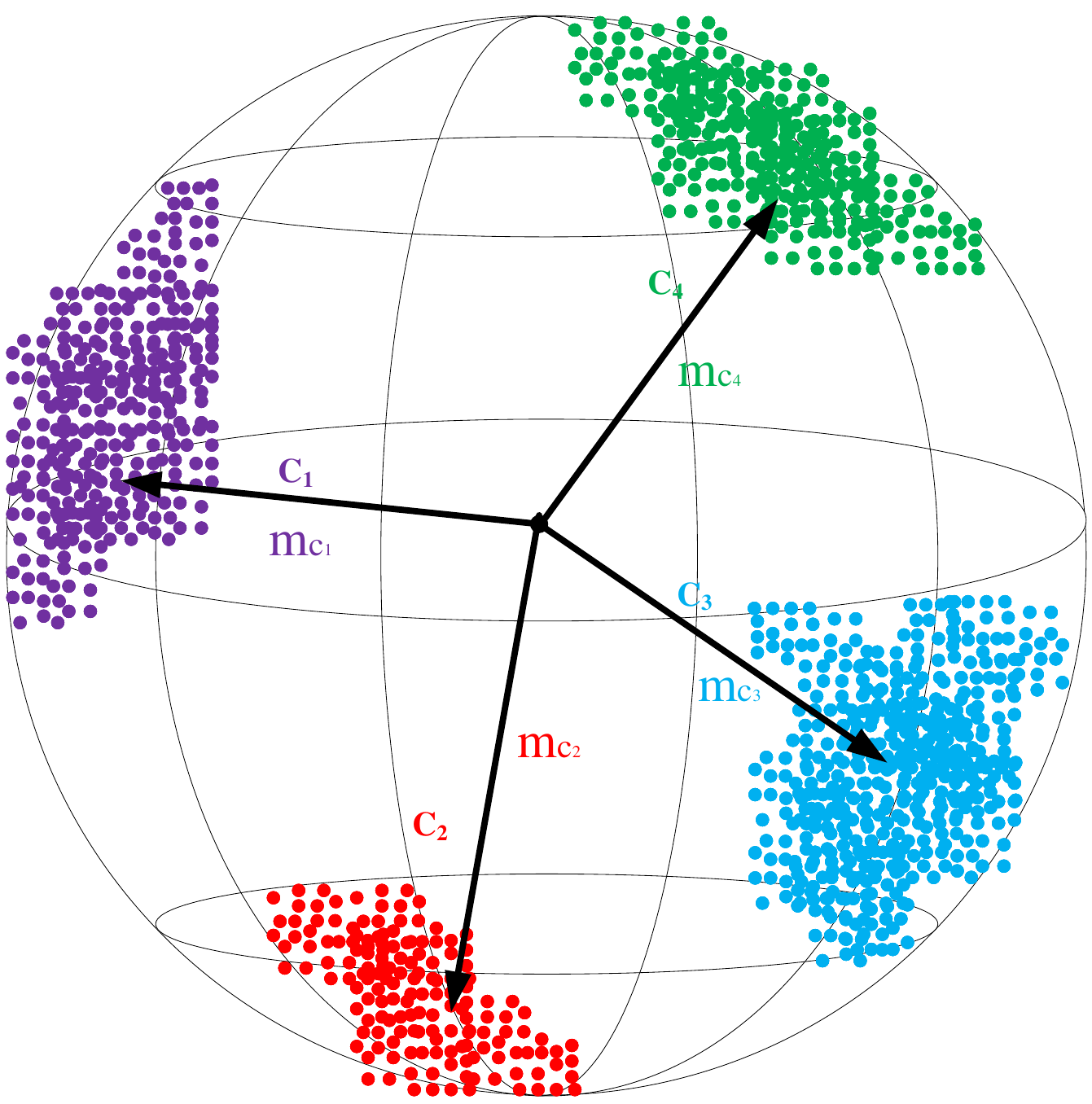}
  \caption{The vFM mixture model of 4 classes. The $\rm{c}_i$ is the $i$-th category, and $\rm{m}_i$ is the mean direction of the $i$-th category.}
  \label{vFMs}
\end{figure}

Moreover, reliability and stability of each layer can stay consistent both for the imperfect-trained and well-trained models. The consistency of sensitiveness can be explained from the vMF distribution that is proposed in \cite{vFM}. We denote the output of a specified layer $l$ as $\mathcal{O}_{l}$, which follows $\mathcal{O}_{l} \in \mathbb{R}^{d}$, where $d$ is the number of filters. Then we will analyze the power of output features through the vMF mixture model, which could assist us to derive the importance of each corresponding convolutional layer. As displayed in Fig. \ref{vFMs}., the output features of a layer can be projected into a unit sphere, and each color stands for one category. Notably, each category $c$ has a mean direction $m_{c}$($c=1,2,3,...,C$) within feature space, and we could use $\cos(\mathcal{O},m_c)$ to indicate the similarity between a specified output feature and category $c$. For the image classification, we usually use softmax to yield the final results of the whole network, $p(y|\mathcal{O}_l)$=Softmax$(\textbf{M}\mathcal{O}_l)$, $\textbf{M}\in\mathbb{R}^{C\times d}$.
Subsequently, the vMF mixture model from \cite{vFM} is introduced to model the radial distribution, which assumes that the likelihood of each feature $\mathcal{O}_l$ belonging to $c\in \{1,2,3,...,C\}$ is constrained by vMF distribution:
\begin{align}
    p(\mathcal{O}_{l})=\sum_{c}p(y=c)p_{vMF}(\mathcal{O}_{l}|y=c),\nonumber \\
    p_{vMF}(\mathcal{O}_{l}|y=c)=C^{d}(\kappa_{l})\exp{[\kappa_{l}\cos(m_c,\mathcal{O}_l)]}.
\end{align}
Where $C^{d}(\kappa_{l})=\frac{(\kappa_{l})^{(d/2-1)}}{(2\pi)^{d/2}I_{(d/2-1)}(\kappa_{l})}$ is a normalization constant for the specified layer, and $I$ is the Bessel function of the first kind at order $d/2-1$. $\kappa_{l}$ determines the variance of $\mathcal{O}_{l}$ orientation $\emph{w.r.t.}$ the mean direction $m_c$, which is associated to the number of filters $d$ \cite{visualizing}. Especially, if the output features of the layer $l$ can be clustered well by the vMF mixture model, the $p(\mathcal{O}_{l})$ will increase because the cosine distance between the $\mathcal{O}_l$ and $m_c$ will be reduced. Then the inference accuracy of the model will increase, and higher importance will be claimed for layer $l$. Namely, the $p(\mathcal{O}_l)$ has positive relation to the representation ability for the original model. Therefore, the $p(\mathcal{O}_l)$ will follow:
\begin{align}
    p(\mathcal{O}_l)\succ Accu,
\end{align}
where "$\succ$" stands for the positive relation, and $Accu$ is the inference accuracy of the model. Therefore, the difference of $p(\mathcal{O}_l)$ will assist us to analyze the importance of each layer for $l=1,2,3,...,\emph{L}$. The relative importance of each layer can be approximated by the inner mutual ratio of $p(\mathcal{O})$, which can be expanded as $\{p(\mathcal{O}_1):p(\mathcal{O}_2):,...,:p(\mathcal{O}_\emph{L})\}$. In this paper, we compare the importance of each layer by analyzing the change of $p(\mathcal{O})$, which can be expanded as $\{\Delta p(\mathcal{O}_1)^{well}:\Delta p(\mathcal{O}_2)^{well}:,...,:\Delta p(\mathcal{O}_\emph{L})^{well}\}$ and $\{\Delta p(\mathcal{O}_1)^{im}:\Delta p(\mathcal{O}_2)^{im}:,...,:\Delta p(\mathcal{O}_\emph{L})^{im}\}$ for the well-trained and imperfect-trained, respectively. When reliability and stability are evaluated, $p(y=c)$ will be same for each layer, and the change of $p_{vMF}$ will be used to analyze above mutual ratio of $\Delta p(\mathcal{O})$.

When we explore the reliability, the feature space of each layer will maintain same as original model, and the $\mathcal{O}_l$ will be diverged because the retraining operation with frozen parameters of layer $l$ will introduce extra noise into the model, which lead to the change of inference accuracy. Accordingly, we denote the variation of the output feature for layer $l$ as $\Delta \mathcal{O}_l^{freeze}$ when the layer $l$ is frozen, then the change of $p(\mathcal{O}_{l})$ both for well-trained and imperfect-trained models can be written as $\Delta p(\mathcal{O}_{l})^{well}$ and $\Delta p(\mathcal{O}_{l})^{im}$, respectively. Because the updating of parameters distribution is smooth for both imperfect-trained and well-trained models when they are only retrained with one epoch, we can get $\{\Delta p(\mathcal{O}_l)^{well}\approx\Delta p(\mathcal{O}_l)^{im}\}$. Consequently, the mutual ratio between the well-trained and imperfect-trained models can be claimed as $\{\Delta p(\mathcal{O}_1)^{well}:\Delta p(\mathcal{O}_2)^{well}:,...,:\Delta p(\mathcal{O}_\emph{L})^{well}\}$ $\approx$ $\{\Delta p(\mathcal{O}_1)^{im}:\Delta p(\mathcal{O}_2)^{im}:,...,:\Delta p(\mathcal{O}_\emph{L})^{im}\}$ when all the layers are explored. Therefore, the evaluation of reliability for each layer will stay stable both for the well-trained and imperfect-trained models.        

Stability is defined as the recovery ability of the structure-sabotaged model when some filters are removed from a specified layer of the original model, which can be calculated by the accuracy gap between original and the structure-sabotaged model that is trained with full epochs. Different with reliability, the removing of filters will impose systematic noise on the network due to the damage of the structure for a specified layer, then the value of concentration parameter $\kappa_{l}$ will be decreased both for the well-trained and imperfect-trained models. We denote the concentration parameters of layer $l$ for original and structure-sabotaged models as $\kappa_{l}^{well}$,$\kappa_{l}^{im}$ and $\kappa_{l}^{well^{'}}$,$\kappa_{l}^{im^{'}}$ with respect to well-trained and imperfect-trained models, respectively. Notably, the difference of structure-sabotaged models for the well-trained and imperfect-trained models is the initialization of parameters across all layers, and similar results can be claimed if the both structure-sabotaged models are retrained fully, namely, $\kappa_{l}^{well^{'}}$ $\approx$ $\kappa_{l}^{im^{'}}$ for layer $l$. Due to the specified imperfect-trained model has skipped the process with rapid updating of parameters, then the difference between $\kappa_{l}^{well}$ and $\kappa_{l}^{im}$ will be small compared with the variation between $\kappa_{l}^{well^{'}}$ and $\kappa_{l}^{well}$, as well as between the $\kappa_{l}^{im^{'}}$ and $\kappa_{l}^{im}$, which means $|\kappa_{l}^{well}-\kappa_{l}^{im}|\ll min(|\kappa_{l}^{well}-\kappa_{l}^{well^{'}}|,|\kappa_{l}^{im}-\kappa_{l}^{im^{'}}|)$. We denote the $|\kappa_{l}^{well}-\kappa_{l}^{well^{'}}|$ and $|\kappa_{l}^{im}-\kappa_{l}^{im^{'}}|$ as $\Delta \kappa_{l}^{well}$ and $\Delta \kappa_{l}^{im}$ for the well-trained and imperfect-trained models, respectively. Then we can claim $\Delta \kappa_{l}^{well}\approx \Delta \kappa_{l}^{im}$, as well as obtaining:
\begin{align}
    \{\Delta \kappa_{1}^{well}:\Delta \kappa_{2}^{well}&:,...,:\Delta \kappa_{\emph{L}}^{well}\}\nonumber\\
    &\approx \nonumber \\
     \{\Delta \kappa_{1}^{im}:\Delta \kappa_{2}^{im}&:,...,:\Delta \kappa_{\emph{L}}^{im}\}\nonumber \\
     &\Downarrow\nonumber \\
      \{\Delta p(\mathcal{O}_{1})^{well}:\Delta p(\mathcal{O}_{2})^{well}&:,...,:\Delta p(\mathcal{O}_{\emph{L}})^{well}\}\nonumber\\
    &\approx \nonumber \\
     \{\Delta p(\mathcal{O}_{1})^{im}:\Delta p(\mathcal{O}_{2})^{im}&:,...,:\Delta p(\mathcal{O}_{\emph{L}})^{im}\}.
\end{align}
Then the evaluation of stability could also maintain stable even when the original model is trained imperfectly. Therefore, both reliability and stability are efficient factors to evaluate the importance of each layer for original model while without considering the effect of individual parameter.
However, when the original model is trained well, reliability may not work well because the accuracy variation is tiny, which can not disentangle the difference of all layers obviously. While stability can perform well because the change of structure for each layer with same magnitude can yield apparently variation of accuracy, then helps us to compare the importance over the whole network. On the contrast, reliability may show superiority when the original model is trained imperfectly. The reliability of the imperfect trained model will lead to blatant change of inference accuracy because the distribution of original model is not optimal, and one epoch retraining with frozen parameters within specified layer can assist us to observe the accuracy variation. But stability may be misleading because the full retraining of each structure-sabotaged model may result in higher fluctuation of accuracy when then parameters distribution of original model is not optimal, then disturb us to distinguish the difference of each convolutional layer. Therefore, we propose the sensitiveness with the sum value of reliability and stability to describe the importance of each layer.
\subsubsection{The Calculation of Sensitiveness}The reliability and stability can be denoted as $f^{r}$ and $f^{s}$ respectively, and the $\emph{P}_\emph{O}$ is denoted as the accuracy of original model. Besides, the accuracy of the parameter-sabotaged model and structure-sabotaged model are denoted as $\emph{P}_{0}$ and $\emph{P}_{max}$ when the structure of original model are kept integral and maximum pruned, respectively. According to the definition of sensitiveness, we can write reliability and stability as:
\begin{align}
\label{preservation_stability}
\centering
    f^{r} = &\emph{P}_0 - \emph{P}_\emph{O}, \nonumber \\
    f^{s} = &\frac{\emph{P}_0 - \emph{P}_{max}}{\lambda \times \emph{R}_{ max}}, \nonumber \\
\end{align}
usually, the value of $f^{r}$ is smaller than $f^{s}$ because the damage of structure will result in more change of representation than the damage of parameters distribution. Therefore, we introduce a scaled constraint $\lambda$ to assure that $f^s$ has same magnitude as $f^r$. Then we could calculate the sensitiveness according to the sum value of $f^{r}$ and $f^{s}$:
\begin{align}
\label{sensitiveness_calculations}
\centering
    \emph{S} = &\gamma f^{r} + (1-\gamma) f^{s}, \nonumber\\
     s.t. \quad  &0<\gamma<1,
\end{align}
\begin{algorithm}[htbp]
	\caption{Hierarchy pruning process}
	\label{hierarchy pruning}
	\begin{algorithmic}[1]
		\Require
		Pruning rounds: $\emph{N}$;
		Layer depth: $\emph{L}$;
		Number of channels in each layer: $\emph{N}_l$;
		Maximum pruning ratio $\emph{R}_{\rm max}$;
		Number of non-zero pruning ratio within each layer: $\emph{T}$.
		\Ensure
		Retrained accuracy of $l$-th layer pruned model with $i$-the pruning epochs during $m$-th pruning round: $a_l^{(i,m)}$;
		Sensitiveness of layer $l$ in the $m$-th pruning round: $\emph{S}_{l}^{m}$;
		Average sensitiveness of each layer: $\emph{S}_l$.
		\State $m=1$.
		\While{$m<\emph{N}$}
		\State $l=1$.
		\If{$l<\emph{L}$}
		\State $r=0$, $i=1$.
		\If{$r<\emph{R}_{\rm max}$}
		\State Remove $r\times \emph{N}_l$ filters from layer $l$ randomly.
		\If{$r==0$}
		\State Retrain parameter-sabotaged model with
		\Statex \quad \quad \quad \quad \quad\;\; frozen parameters for the $l$-th layer and \Statex \quad \quad \quad \quad \quad\;\; yield $a_l^{i,m}$.
		\Else
		\State Retrain the structure-sabotaged model and \Statex \quad \quad \quad \quad \quad\;\; yield $a_l^{i,m}$.
		\EndIf
		\State $r = r +i\frac{\emph{R}_{max}}{\emph{T}}$.
		\Else
		\State Calculate $\emph{S}^{m}_{l}$ with $a_{l}^{(i,m)}$ according to the \Statex \quad \quad \quad \quad \; Equation (\ref{sensitiveness_calculations}).
		\State $l++$.
		\EndIf
		\Else
		\State $m++$.
		\EndIf
		\EndWhile
		\State Return $\emph{S}_l$ with Equation (\ref{average_sensitiveness}).
	\end{algorithmic}
\end{algorithm}
where $\gamma$ is the control factor which allocates different weights to $f^{r}$ and $f^{s}$, and $\emph{S}$ stands for the score of sensitiveness. In order to promote convincing results of $\emph{S}$, more weights should be authorized to reliability because the fluctuation of stability is heavier, and lower weights for stability can reduce the fluctuation of $\emph{S}$. Besides, reliability is more credible to reflect the importance of each layer because the interference of extra noise is tiny compared with the systematic noise for stability. Consequently, we will assign more weights for reliability and relative lower weights for stability.
\subsubsection{Hierarchy Pruning}
Hierarchy pruning means the sensitiveness of all layers can be explored in parallel due to each layer can be regarded as a individual item, namely, when a layer is specified to be pruned, other layers will stay unchanged. Notably, when we measure the  stability, a predefined pruning ratio $\emph{R}_{max}$ should be set in advance to make sure the structure of the specified layer is damaged with sufficient efficiency. When $\emph{R}_{max}$ is set with a tiny value, it will be easy to recovery baseline accuracy only with few training epochs for each structure-sabotaged model, then similar $f^{s}$ will be yielded for each layer, which disturb us to obtain the correct evaluation of stability. On the contrast, if $\emph{R}_{max}$ is set too large, the filters within the specified layer will be less than the minimum boundary to support the representation ability of the structure-sabotaged model, then the retraining operation can not instruct the pruned model to be optimized along the right dimension, which twists the evaluation of stability for each layer. Therefore, $\emph{R}_{max}$ should be considered carefully. Due to the random removing of filters during the evaluation process, heavier fluctuation may appear to impede us to explore the stability. Therefore, we will remove filters progressively to reduce the fluctuation. Specifically, we define a pruning ratio set from zero to $\emph{R}_{max}$ with uniform distribution:
\begin{align}
    \textbf{R}=Ascending&\{0,\emph{R}_1,\emph{R}_2,...,\emph{R}_{max}\},\nonumber \\
    s.t. \quad &num(\textbf{R})=\emph{T}+1, \nonumber \\
    &\emph{R}_n-\emph{R}_{n-1}=\frac{\emph{R}_{max}}{\emph{T}},\nonumber \\
    &n\leq max,
    \label{pruning_set}
\end{align}
where $num(\cdot)$ will count the number of elements for the specified set, and $\emph{T}$ stands for the number of pruning samples for each layer. Then there will be $\frac{\emph{R}_{max}}{\emph{T}}$ percent filters being removed from the specified layer after each pruning. Namely, the next filters removal will be based on previous structure-sabotaged model, and the pruning process will be ended until the pruning ratio reaches to the predefined $\emph{R}_{max}$. Correspondingly, the accuracy of each structure-sabotaged model can be written as the set $\textbf{P}$:
\begin{align}
    \textbf{P}=\{\emph{P}_{0},\emph{P}_1,...,\emph{P}_{max}\}.
    \label{performance}
\end{align}
The progressive pruning strategy can help us to observe the recovering of the accuracy for each pruning sample, then we can judge whether the results of $\emph{P}_{max}$ is credible according to the $\textbf{P}$ during whole pruning process. If the elements within $\textbf{P}$ are not smooth, the $\emph{P}_{max}$ may be incredible because the random removing of filters may result in heavier fluctuation, which disturb us to obtain the real evaluation of importance for the specified layer. Therefore, we repeat each hierarchy pruning process with several rounds and pick up the smooth pruning results to decrease the effect of random pruning, then yield credible evaluation of sensitiveness for each layer. The overall pruning results can be written as following matrix:
\begin{align}
&\textbf{P}_{l}^{N}=
\begin{bmatrix}
\bm{p}_1^1,\bm{p}_1^2,...,\bm{p}_1^\emph{N}\\
\bm{p}_2^1,\bm{p}_2^2,...,\bm{p}_2^\emph{N}\\
...\\
\bm{p}_l^1,\bm{p}_l^2,...,\bm{p}_l^\emph{N}  \\
\end{bmatrix}\nonumber \\
&s.t. \quad \bm{p}_{l}^{\emph{N}}=[a_l^{(1,\emph{N})},a_l^{(2,\emph{N})},...,a_l^{(\emph{T},\emph{N})}],
\end{align}
where $\emph{N}$ is the number of pruning rounds, and $l$ is the index of convolutional layer. The $\bm{p}_l^{\emph{N}}$ stands for a row vector that contains a series of retrained accuracy for $l$-th-layer structure-sabotaged models, and $a_{l}^{(\emph{T},\emph{N})}$ stands for the $\emph{T}$-th retrained accuracy within $\bm{p}_l^{\emph{N}}$, which satisfies $\bm{a}_l^{(\emph{T},\emph{N})}=\emph{P}_{l}^{(\emph{T},\emph{N})}$ according to Equation (\ref{performance}). Then we could flatten the retrained accuracy by averaging the results over all pruning rounds while excluding the fluctuated samples. In this paper, Pearson correlation coefficient analysis could be adopted to abandon the fluctuated pruning rounds when hierarchy pruning is repeated. At first, we will calculate the variance of retrained accuracy for the $l$-th-layer structure-sabotaged model during each hierarchy pruning round:
\begin{align}
\sigma_{(l,\emph{N})}^{2}=\frac{\sum\limits_{i=1}^{\emph{T}}(a_{l}^{(i,\emph{N})}-\frac{\sum\limits_{i=1}^{\emph{T}}a_{l}^{(i,\emph{N})}}{\emph{T}})}{\emph{T}},
\end{align}
where $\sigma_{(l,\emph{N})}^{2}$ is the variance of $\bm{p}_{l}^{\emph{N}}$. Then we can locate the index of the most flat pruning round for $l$-th layer:
\begin{align}
I=\arg\min\limits_{m}\sigma_{(l,m)}^{2},\nonumber \\
s.t. \quad m=1,2,...,\emph{N},
\end{align}
where $I$ is the index of the most flat pruning round for layer $l$. Then we can compare the $I$-th pruning round with others through Pearson correlation coefficient:
\begin{footnotesize}
\begin{align}
 \rho^{(m,I)}_{l} = \frac{\sum\limits_{i=1}^{T}a_l^{(i,m)}a_{l}^{(i,I)}-\frac{\sum\limits_{i=1}^{T}a_l^{(i,m)}\sum\limits_{i=1}^{T}a_l^{(i,I)}}{T}}{\sqrt{(\sum\limits_{i=1}^{T}{(a_{l}^{(i,m)})^2}-\frac{\sum\limits_{i=1}^{T}(a_{l}^{(i,m)})^2}{T})(\sum\limits_{i=1}^{T}(a_{l}^{(i,I)})^2-\frac{\sum\limits_{i=1}^{T}(a_{l}^{(i,I)})^2}{T})}},\nonumber\\
 s.t. \quad m\leq N, m \neq I,
\end{align}
\end{footnotesize}
where $\rho^{(m,I)}_{l}$ is the Pearson correlation coefficient between $m$-th and $I$-th pruning round for layer $l$. Then we will average the pruning results of relative flat pruning rounds while abandoning fluctuated ones according to $\rho^{(m,I)}_{l}$. The final sensitiveness of layer $l$ can be calculated as:
\begin{align}
\emph{S}_l = \{\frac{1}{\emph{N}^{'}}\sum\limits_{j}\emph{S}_l^j|\rho^{(j,I)}_{l}>0.6\},\nonumber \\
s.t. \quad \emph{N}^{'}\leq \emph{N},
\label{average_sensitiveness}
\end{align}
where $\emph{N}^{'}$ is the total pruning rounds if $\rho^{(j,I)}_{l}>0.6$, which means the results of $j$-th and $I$-th pruning round are highly correlated. Then the final $\emph{S}_l$ will contribute to yield reliable evaluation for layer $l$. The overview of hierarchy pruning process is concluded as Algorithm-\ref{hierarchy pruning}.
\subsection{Layer Grouping with Specified Sensitiveness}
After the obtainment of sensitiveness, we should assign each convolutional layer of the original model with different pruning ratio according to the value of $\emph{S}$. However, the allotting process may be cumbersome when the original model is large. Therefore, layers could be classified into different groups if they show similar value of $\emph{S}$, namely, layers with similar sensitiveness will be slimmed with same pruning ratio in the final pruning process. The grouping operation can be describe as:
\begin{align}
\label{layers_grouping}
\textbf{G} =\{\textbf{G}_k&|\emph{S}_i\approx \emph{S}_j,...,\approx \emph{S}_o\},\nonumber\\
s.t. \quad &k\leq \emph{K} \leq \emph{L},\nonumber \\
     &\{i,j,...,o\}\in \{1,2,...,\emph{L}\},\nonumber \\
     &\textbf{G}_k=\{\bm{f}_i,\bm{f}_j,...,\bm{f}_o\}.
\end{align}
\begin{algorithm}[htbp]
	\caption{Iterative pruning process}
	\label{iterative-pruning}
	\begin{algorithmic}[1]
		\Require
		Average sensitiveness of each layer that is calculated from Algorithm-\ref{hierarchy pruning}: $\emph{S}_l$; Number of layers-group: $\emph{K}$;
		Pruning ratio in the $k$-th group: $\emph{R}_{g}^{k}$.
		\Ensure
		Retrained accuracy of the final pruned model: $\emph{P}_{r}$; Best retrained accuracy: $\emph{P}^{\rm best}_{r}$.
		\State Divide layers into $\emph{K}$ groups by descending order with the value of $\emph{S}_l$.
		\State set $k=1$.
		\For{$k<\emph{K}$}
		 \State $P^{\rm best}_{r}=0$.
		\State Remove filters from each layer that belongs to $k$-th
		\Statex \quad \;group with the pruning ratio $\emph{R}_{g}^{k}$.
		\State Retrain pruned model with free gradient for all layers
		\Statex \quad \; and yield $\emph{P}_{r}.$
		\If{$\emph{P}^{\rm best}_{r}$ $<$ $\emph{P}_{r}$ }
		\State Update $P^{\rm best}_{r}$=$\emph{P}_{r}$.
		\State Increase pruning ratio $\emph{R}_{g}^{k}$.
		\Else
		\State $k++$.
		\EndIf
		\EndFor
		\State Return $\emph{P}^{\rm best}_{r}$.
	\end{algorithmic}
\end{algorithm}
Where $\textbf{G}_k$ represents for the $k$-th group which contains the layers that share similar sensitiveness, and $\emph{K}$ is the number of groups. Notably, each layer only can be classified into one group to avoid the multi-pruning on one layer. Then the layers within same group will be specified with same pruning ratio, which could speed up final pruning process efficiently.

\subsection{Iterative Pruning Process}
Honestly, exact slimming for original model only with one time is hard. Therefore, it is better to resort iterative strategy to search for an appropriate overall pruning ratio across all layers. The details about the iterative pruning process are concluded as Algorithm \ref{iterative-pruning}. In the step 1, we divide convolutional layers of the original model into $\emph{K}$ groups according to the value of $\emph{S}$ with ascending order, which means the sensitiveness of layers within the groups will increase with the growing of group index. When the group index is less than $\emph{K}$, we will initialize the best retrained accuracy of the pruned model $\emph{P}_r^{best}$ as 0 in step 4. Then in the step 5, we will remove filters from $k$-th group with the pruning ratio $\emph{R}_{g}^{k}$, and consequently, the pruned model will be retrained with free gradient to yield the accuracy $\emph{P}_r$. For the $k$-th group, we will update $\emph{P}_r^{best}$ with $\emph{P}_r$, as well as increasing the pruning ratio $\emph{R}_{g}^{k}$ in step 8 if the $\emph{P}_r^{best}<\emph{P}_r$, otherwise, we will update the group index $k$ as $k+1$. The pruning will be ceased until $k>\emph{K}$.

In order to illustrate the efficiency of our pruning framework, both pure chain structure and skip structure models are pruned across different datasets in following parts.
\section{Experiments}
\begin{table*}[htbp]
   		\small
      \centering
      \caption{Baseline And Pruning Performance For Different Fully Trained Models On Various Datasets.}
      \setlength{\tabcolsep}{4mm}{
        \begin{tabular}{lccccc}
        \multicolumn{6}{c}{(a) Results on CIFAR-10} \\
        \toprule
        Model & Accuracy & FLOPs (MACs) & Pruned & Parameters & Pruned \\
        \midrule
        VGG-16(baseline) & 0.92  & $3.1\times 10^8$  &-- --& 14.98M &-- -- \\
        VGG-16(60\% Pruned) & 0.925 & $2.1\times 10^8$   & 32.26\% & 2.60M & 82.60\% \\
        VGG-16(65\% Pruned) & 0.92  & $1.9\times 10^8$   & 38.71\% & 2.15M & 85.65\% \\
        \toprule
        \\
        \multicolumn{6}{c}{(b) Results on MNIST} \\
        \toprule
        Model & Accuracy & FLOPs (MACs) & Pruned & Parameters & Pruned \\
        \midrule
        Conv-4(baseline) & 0.99  & $3.2\times10^8$  &-- --& 14.4M & -- -- \\
        Conv-4(80\% Pruned) & 0.99  & $1\times 10^7$     & 96.87\% & 1.37M & 90.49\% \\
        Conv-4(95\% Pruned) & 0.98  & $2\times 10^6$     &  99.38\%     & 0.66M & 95.42\% \\
        \toprule
        \\
        \multicolumn{6}{c}{(c) Results on CIFAR-100} \\
        \toprule
        Model & Accuracy & FLOPs (MACs) & Pruned & Parameters & Pruned \\
        \midrule
        ResNet-18(baseline) & 0.851  & $4\times 10^7$  &-- --& 11.68M &-- --  \\
        ResNet-18(35\% Pruned) & 0.857 & $2\times 10^7$   & 50\% & 2.22M & 81.00\% \\
        ResNet-18(37\% Pruned) & 0.851  & $2\times 10^7$   & 50\% & 2.02M & 82.79\% \\
        ResNet-18(40\% Pruned) & 0.848  & $2\times 10^7$   & 50\% & 1.76M & 84.86\% \\
        \bottomrule
        \end{tabular}}%
    \label{benchmarks_pruning}%
  \end{table*}%
\renewcommand\arraystretch{1.3}
    \begin{table*}[htbp]
	\footnotesize
	\centering
	\caption{Scores Of Each Convolutional Layer For Different Fully Trained Models On Isolated Datasets.}
	\begin{minipage}[htbp]{150mm}
		\centering
		\setlength{\tabcolsep}{6.0mm}{
			\begin{tabular}{|p{2.5cm}|p{0.9cm}|p{0.15cm}|p{1.1cm}|p{1.1cm}|p{1.0cm}|}
				\multicolumn{6}{c}{(a) Scores of sensitiveness for VGG-16 on CIFAR-10} \\
				\hline
				Models (datasets) & Layers & $\gamma$   & {$f^r$} & {$f^s$} & {$\emph{S}$} \\
				\hline
				\multirow{13}[0]{*}{VGG-16 (CIFAR-10)} & Conv1 & \multirow{13}[0]{*}{$\frac{2}{3}$}  & 5.1E-03 & 4.0E-03 & 4.7E-03 \\
				& Conv2 &              & 6.6E-03 & 7.4E-03 & 6.9E-02 \\
				& Conv3 &              & 5.6E-03 & 4.3E-03 & 5.2E-03 \\
				& Conv4 &            & 5.2E-03 & 5.2E-03 & 5.2E-03 \\
				& Conv5 &              & 4.1E-03 & 3.3E-03 & 3.8E-03 \\
				& Conv6 &             & 3.8E-03 & 3.4E-03 & 3.7E-03 \\
				& Conv7 &         & 3.7E-03 & 3.4E-03 & 3.6E-03 \\
				& Conv8 &             & 3.4E-03 & 2.9E-03 & 3.2E-03 \\
				& Conv9 &             & 3.0E-03 & 1.3E-03 & 2.4E-03 \\
				& Conv10 &            & 2.9E-03 & 1.7E-03 & 2.5E-03 \\
				& Conv11 &            & 1.5E-03 &6.8E-04 & 1.2E-03\\
				& Conv12 &            & 1.1E-03 &3.4E-04 & 8.5E-04 \\
				& Conv13 &            & 8.0E-04 & 2.4E-04 & 1.3E-04 \\
				\hline
			\end{tabular}
		}
	\end{minipage}
	
	\begin{minipage}[htbp]{150mm}
		\centering
		\setlength{\tabcolsep}{6.0mm}{
			\begin{tabular}{|p{2.5cm}|p{0.9cm}|p{0.15cm}|p{1.1cm}|p{1.1cm}|p{1.0cm}|}
				\multicolumn{6}{c}{(b) Scores of sensitiveness for Conv-4 on MNIST} \\
				\hline
				Models (datasets) & Layers & $\gamma$  &$f^r$  & $f^s$ & $\emph{S}$ \\
				\hline
				\multirow{4}[0]{*}{Conv-4 (MNIST)} & Conv1 & \multirow{4}[0]{*}{$\frac{2}{3}$}  & -1.0E-04 & 1.0E-02 & 1.0E-02 \\
				& Conv2 &              & -7.0E-04 & 2.2E-03 & 1.5E-03 \\
				& Conv3 &              & -1.2E-03 & 9.9E-04 & -2.1E-04 \\
				& Conv4 &              & -7.0E-04 & 2.5E-04 & -4.5E-04 \\
				\hline
			\end{tabular}
		}
	\end{minipage}
	
	\begin{minipage}[htbp]{150mm}
		\centering
		\setlength{\tabcolsep}{6.0mm}{
			\begin{tabular}{|p{2.5cm}|p{0.9cm}|p{0.15cm}|p{1.1cm}|p{1.1cm}|p{1.0cm}|}
				\multicolumn{6}{c}{(c) Scores of sensitiveness for ResNet-18 on CIFAR-100} \\
				\hline
				Models (datasets) & Layers & $\gamma$ & $f^r$ & $f^s$ & $\emph{S}$ \\
				\hline
				\multirow{8}[0]{*}{ResNet-18 (CIFAR-100)} & Conv1 & \multirow{8}[0]{*}{$\frac{2}{3}$}  & 2.2E-03 & 4.3E-03 & 2.9E-03 \\
				& Conv2 &          & 2.E-03 & 2.1E-03 & 2.4E-03 \\
				& Conv3 &              & 2.4E-03 & 8.5E-03 & 4.4E-03 \\
				& Conv4 &              & 2.8E-03 & 3.3E-03 & 3.0E-03 \\
				& Conv5 &              & 3.3E-03 & 2.8E-03 & 3.1E-03 \\
				& Conv6 &              & 8.0E-04 & 2.7E-04 & 6.2E-04 \\
				& Conv7 &              & 1.9E-03 & 9.7E-05 & 1.3E-03 \\
				& Conv8 &              & 1.4E-03 & 2.7E-04 & 1.0E-03 \\
				\hline
			\end{tabular}
		}
	\end{minipage}
	\label{tab:vgg16score}%
\end{table*}
In this section, we apply the sensitiveness based pruning framework on pure chain structure networks with relative deep architecture (VGG-16) and shallow architecture (customized-Conv-4) with CIFAR-10 and MNIST, as well as on skip structure (ResNet-18) with CIFAR-100. At first, original networks were trained with different epochs. For VGG-16, the training epochs were set to 50, 100 and 150, respectively. For Conv-4, we trained the original model with 5, 10 and 20 epochs, respectively. Finally, ResNet-18 was trained with 50, 100 and 150 epochs, respectively. Notably, there was a hyper-parameter $\gamma$ instructing $\emph{S}$, which imposed huge affect on the assessment of sensitiveness. Analyzed in section \ref{methodology}, reliability should be prioritized, so we gave $\gamma$ with the value $\frac{2}{3}$. At the same time, $\lambda$ was set to 10 for assuring appropriate magnitude for stability in this paper, and test round $N$ was set to 10 for reducing of fluctuation during hierarchy pruning process.

Besides, the maximum pruning ratio $\emph{R}_{max}$ for each layer was set as 0.96, and the retraining epochs for the structure-sabotaged models were set as 20, 10, 30 for VGG-16, Conv-4 and ResNet-18, respectively. During final iterative pruning process, the retrained epochs were revised to 100 for VGG-16 and ResNet-18, and Conv-4 would be retrained with 20 epochs. Simultaneously, the initial parameters of final pruned models were inherited from original network to accelerate retraining process. The benchmarks with fully training and pruning results are illustrated on Table \ref{benchmarks_pruning}. Obviously, VGG-16 and Conv-4 shows better pruning results than ResNet-18, and the details about the pruning experiments are discussed in following parts.
  \subsection{VGG-16 on CIFAR-10}
\begin{figure*}[htbp]
      \centering
      \subfigure[Untrained model (50 epochs).]{
        \begin{minipage}[t]{0.3\linewidth}
          \centering
          \includegraphics[width=2.0in]{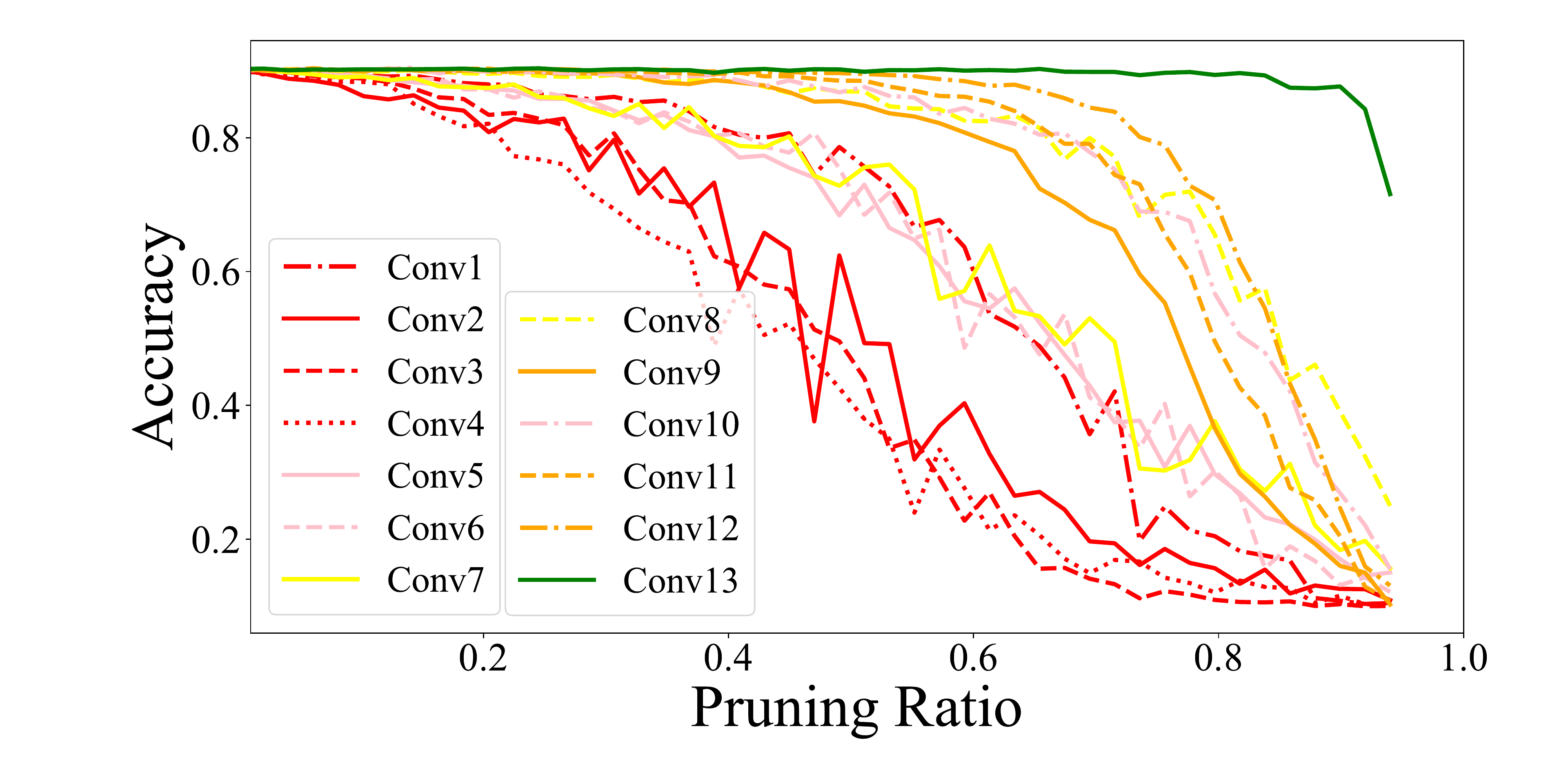}
        \end{minipage}
        }
      \subfigure[Untrained model (100 epochs).]{
        \begin{minipage}[t]{0.3\linewidth}
          \centering
          \includegraphics[width=2.0in]{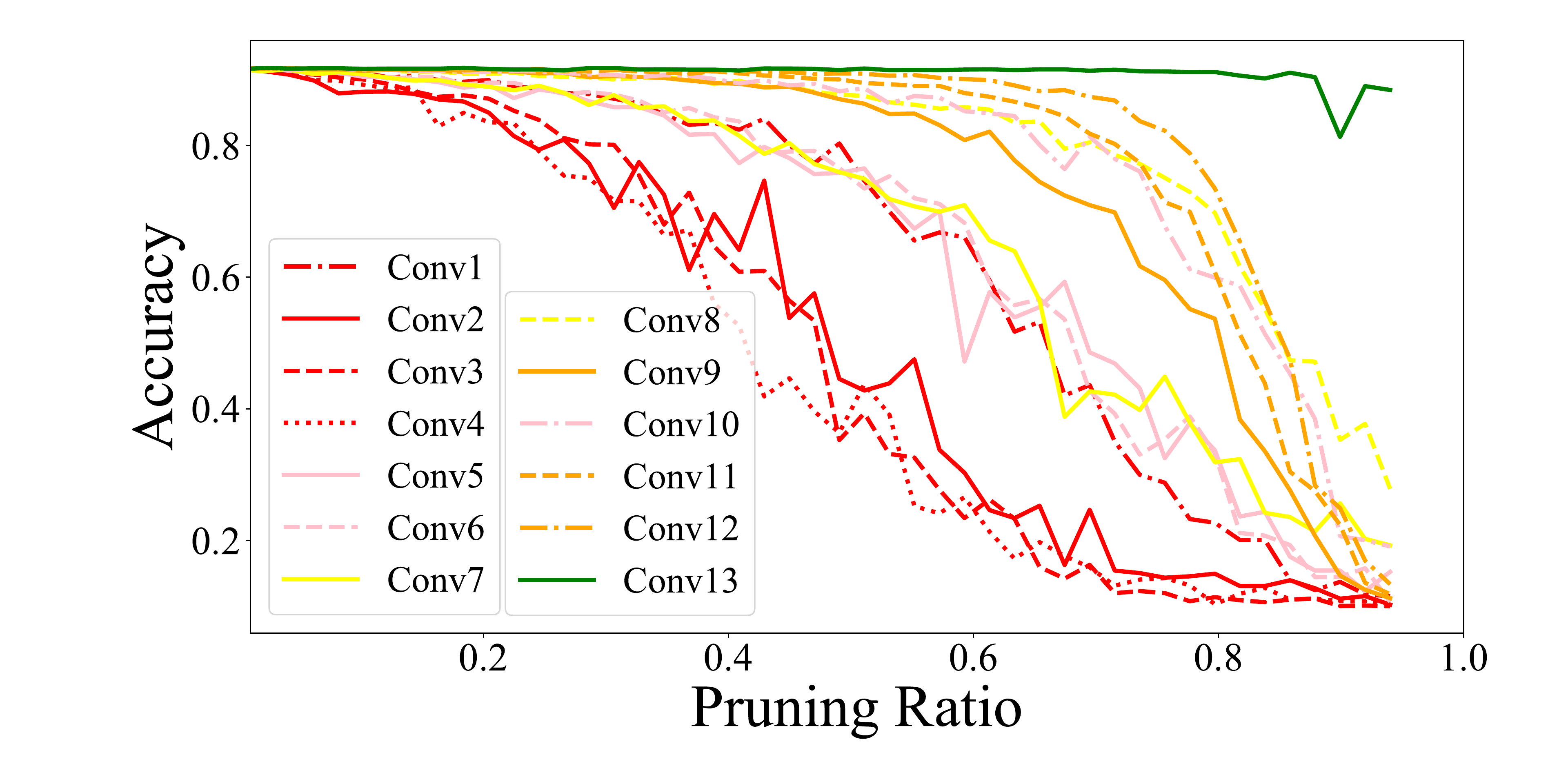}
        \end{minipage}
      }
         \subfigure[Untrained model (150 epochs).]{
        \begin{minipage}[t]{0.3\linewidth}
          \centering
          \includegraphics[width=2.0in]{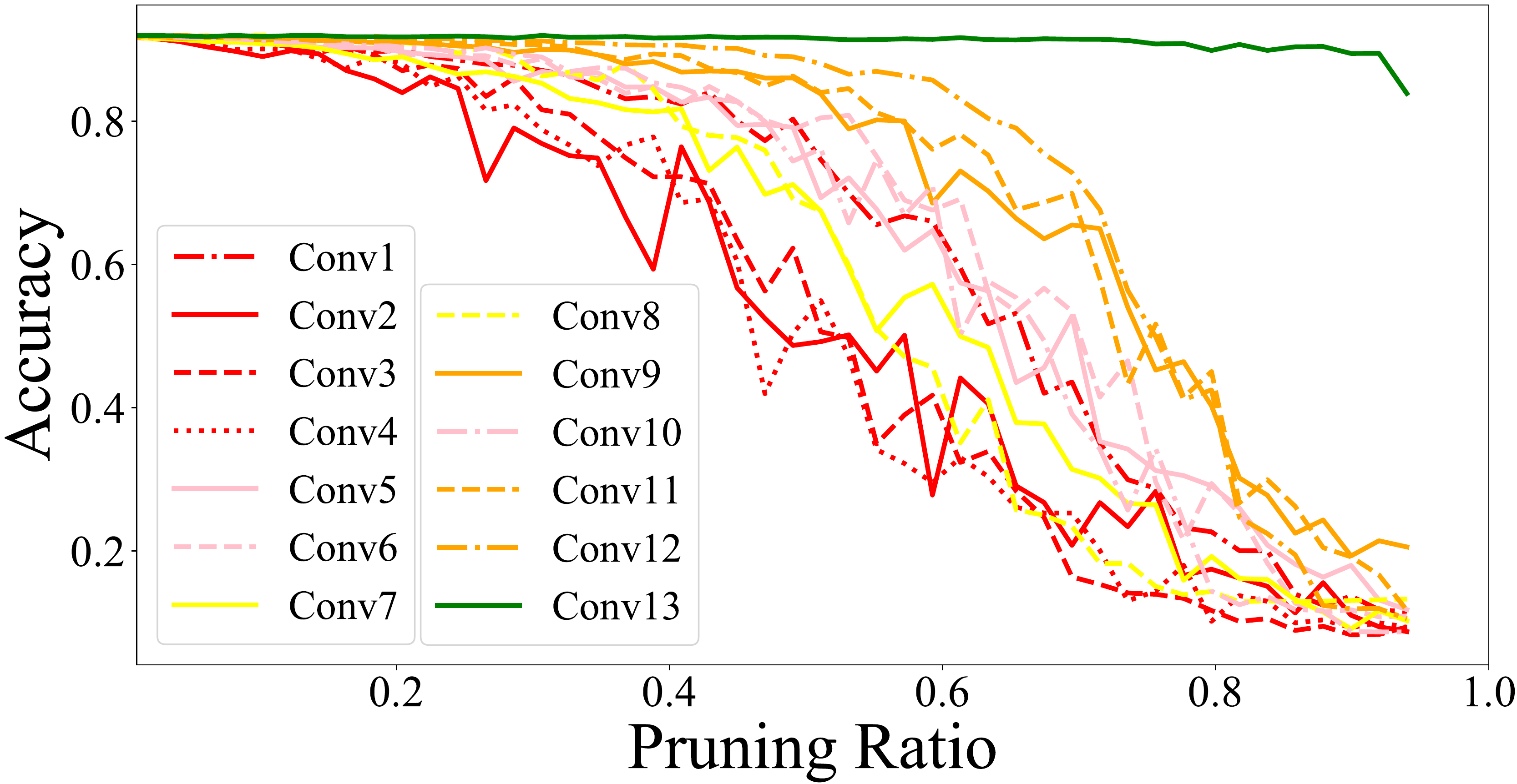}
        \end{minipage}
        }

      \subfigure[Retrained model (50 epochs).]{
        \begin{minipage}[t]{0.3\linewidth}
          \centering
          \includegraphics[width=2.0in]{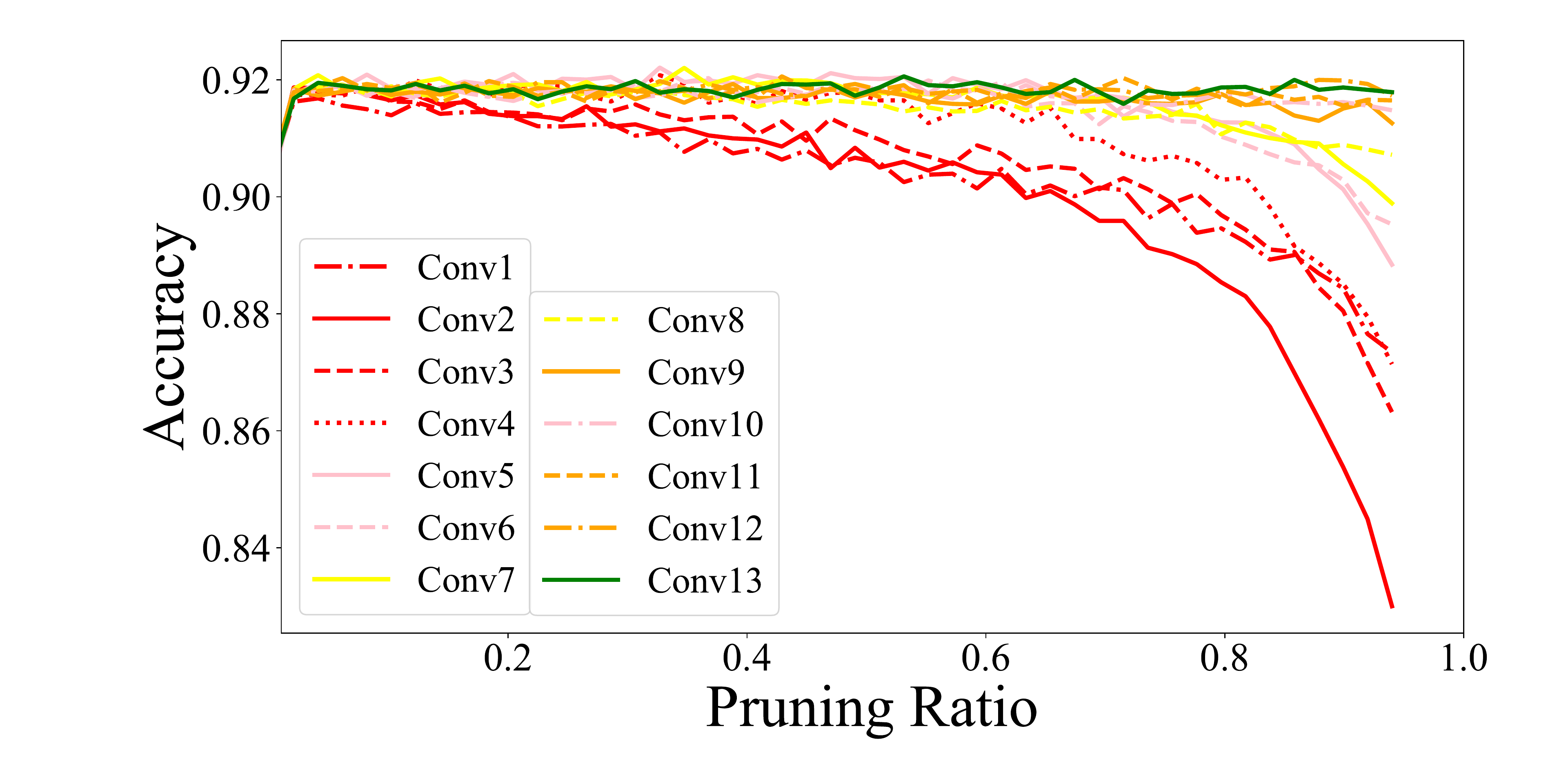}
        \end{minipage}
      }
         \subfigure[Retrained model (100 epochs).]{
        \begin{minipage}[t]{0.3\linewidth}
          \centering
          \includegraphics[width=2.0in]{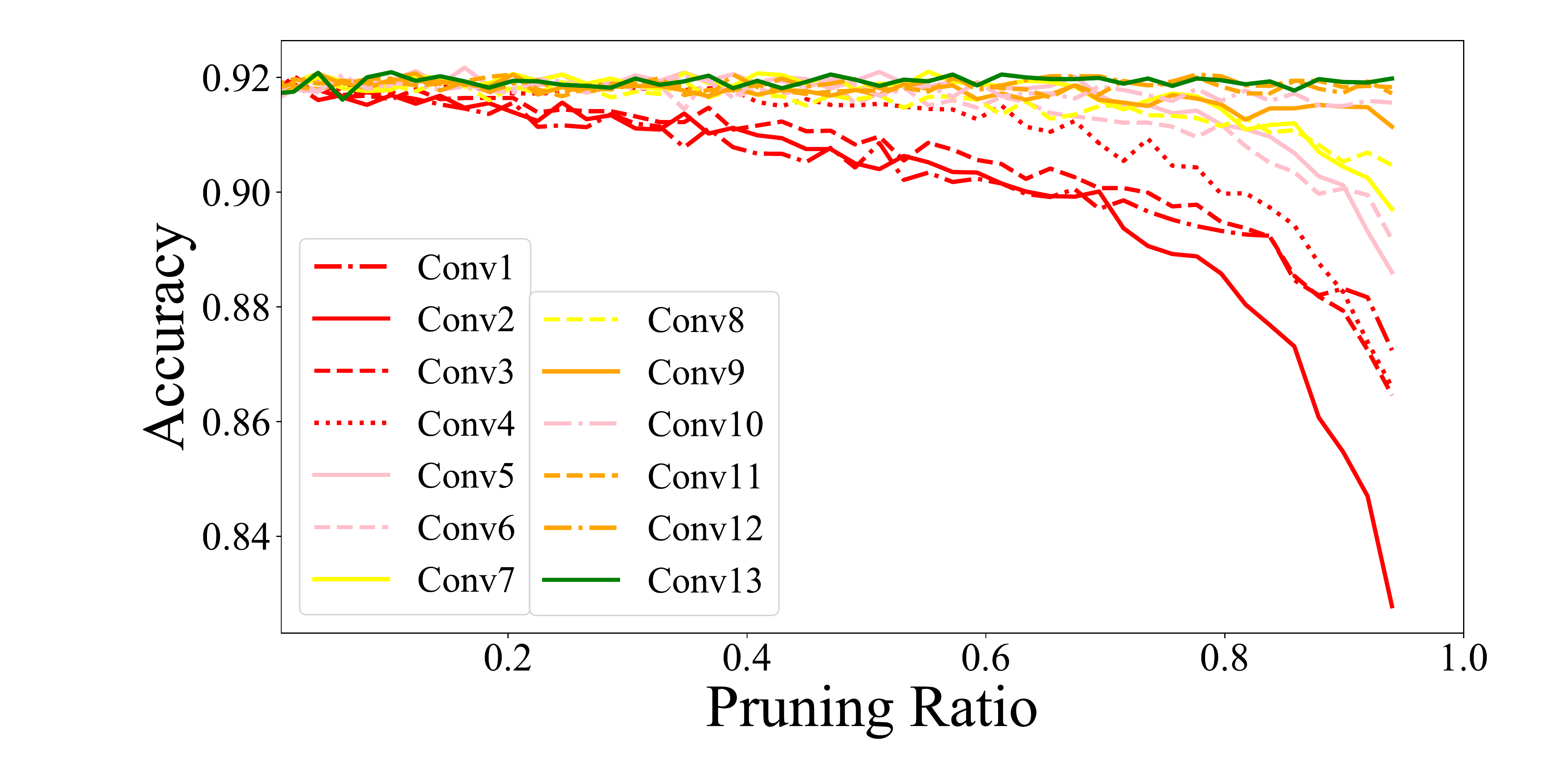}
        \end{minipage}
        }
      \subfigure[Retrained model (150 epochs).]{
        \begin{minipage}[t]{0.3\linewidth}
          \centering
          \includegraphics[width=2.0in]{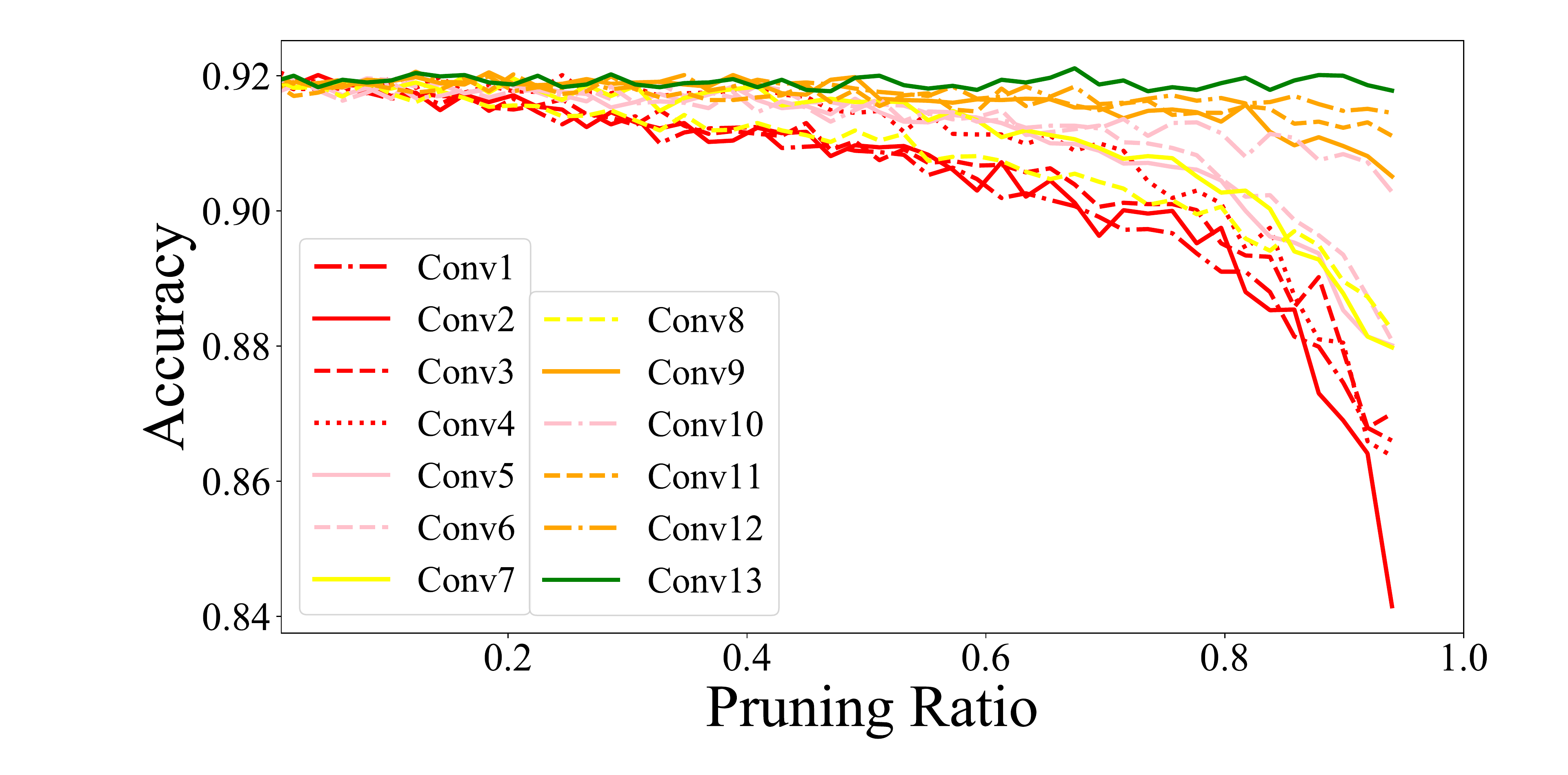}
        \end{minipage}
      }
      \caption{The performance of hierarchy pruning for VGG-16 on CIFAR-10 when original models are trained differently.}
      \label{tab:vgg16pruning}
    \end{figure*}

\begin{figure*}[htbp]
  \centering
  \subfigure[]{
    \begin{minipage}[t]{0.32\linewidth}
      \centering
      \includegraphics[width=2.0in]{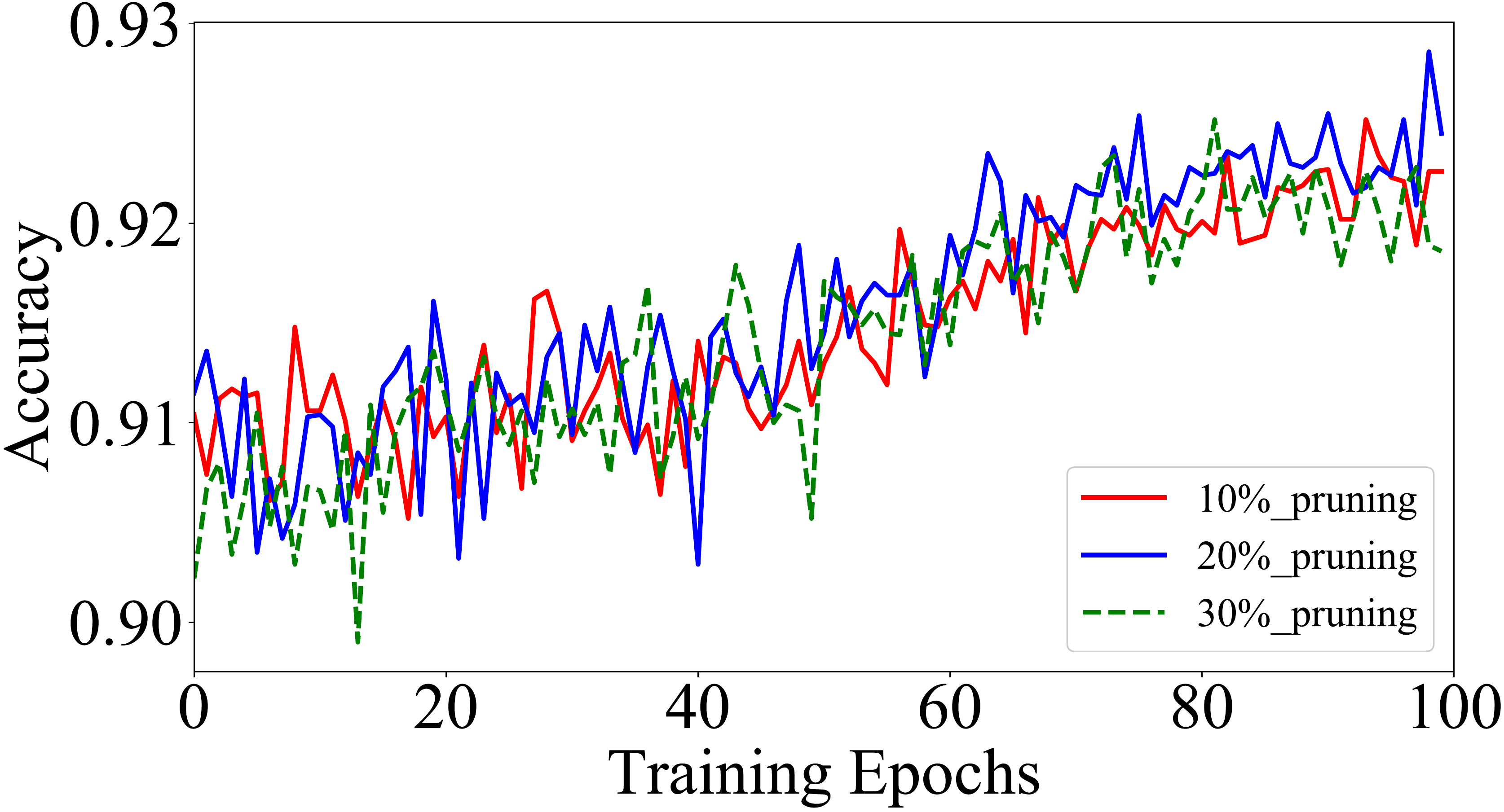}
    \end{minipage}
  }
  \subfigure[]{
    \begin{minipage}[t]{0.32\linewidth}
      \centering
      \includegraphics[width=2.0in]{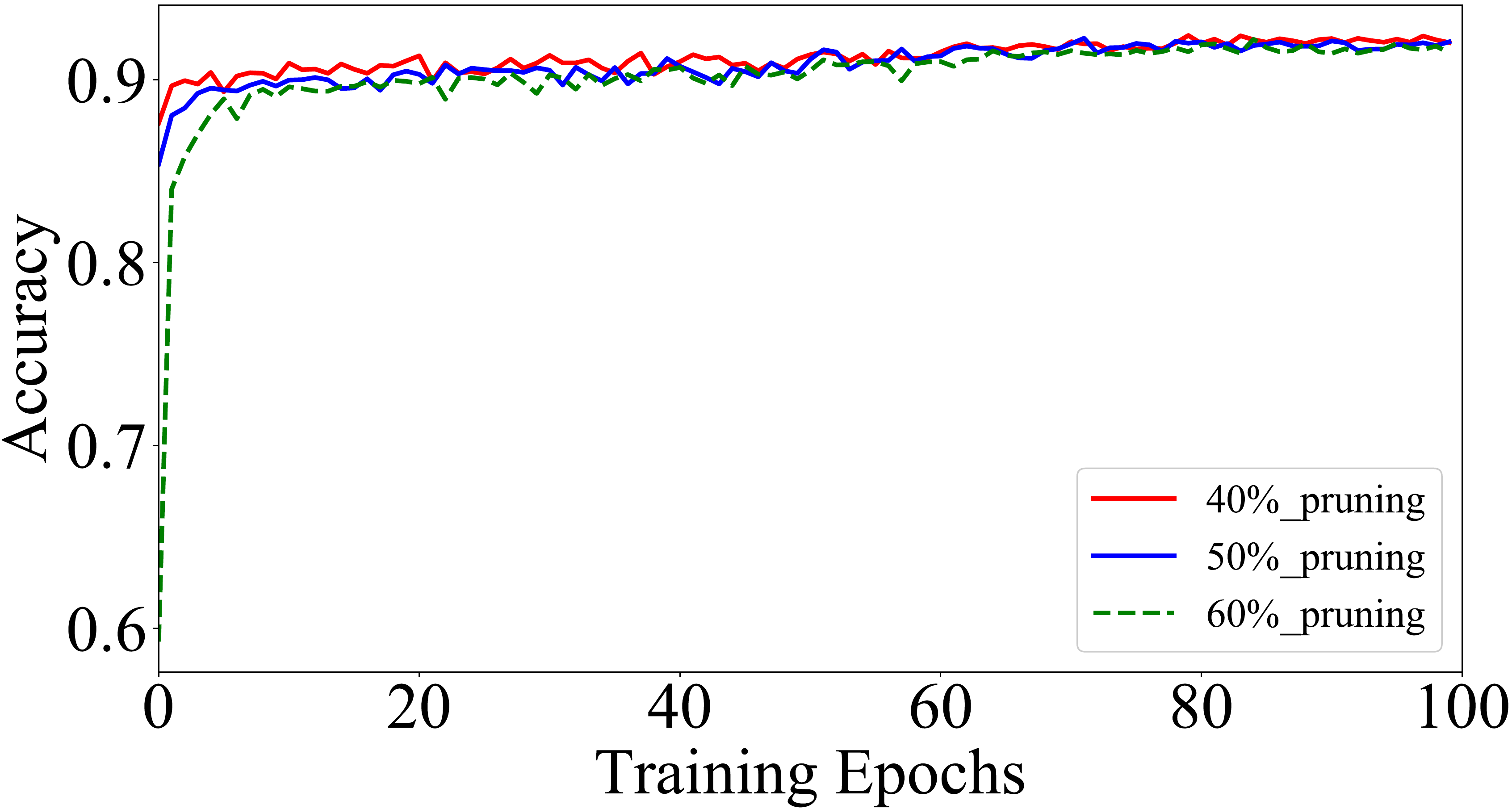}
    \end{minipage}%
  }%
  \subfigure[]{
    \begin{minipage}[t]{0.32\linewidth}
      \centering
      \includegraphics[width=2.0in]{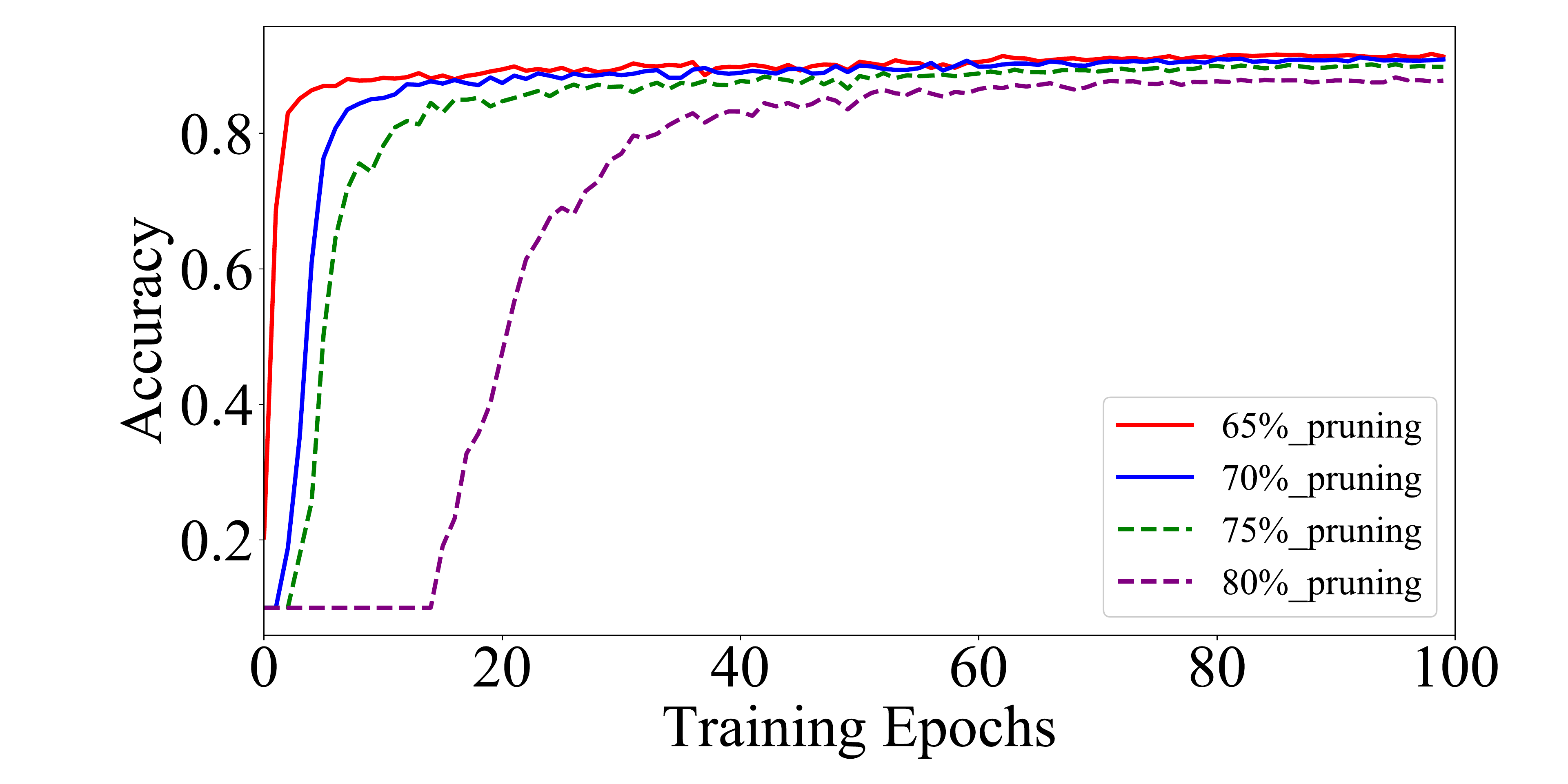}
  \end{minipage}%
  }%
  \caption{Retrained results for VGG-16 on CIFAR-10 with iterative pruning. In (a), the pruning ratio is set to 10\%, 20\% and 30\%. In (b), the pruning ratio is set to to 40\%, 50\%, 60\% and in (c), the pruning ratio is set to 65\%, 70\%, 75\%, 80\%, respectively.}
  \label{tab:vgg16retraining}
\end{figure*}

\begin{figure*}[htb]
      \centering
      \subfigure[]{
        \begin{minipage}[t]{0.48\linewidth}
          \centering
          \includegraphics[width=2.8in]{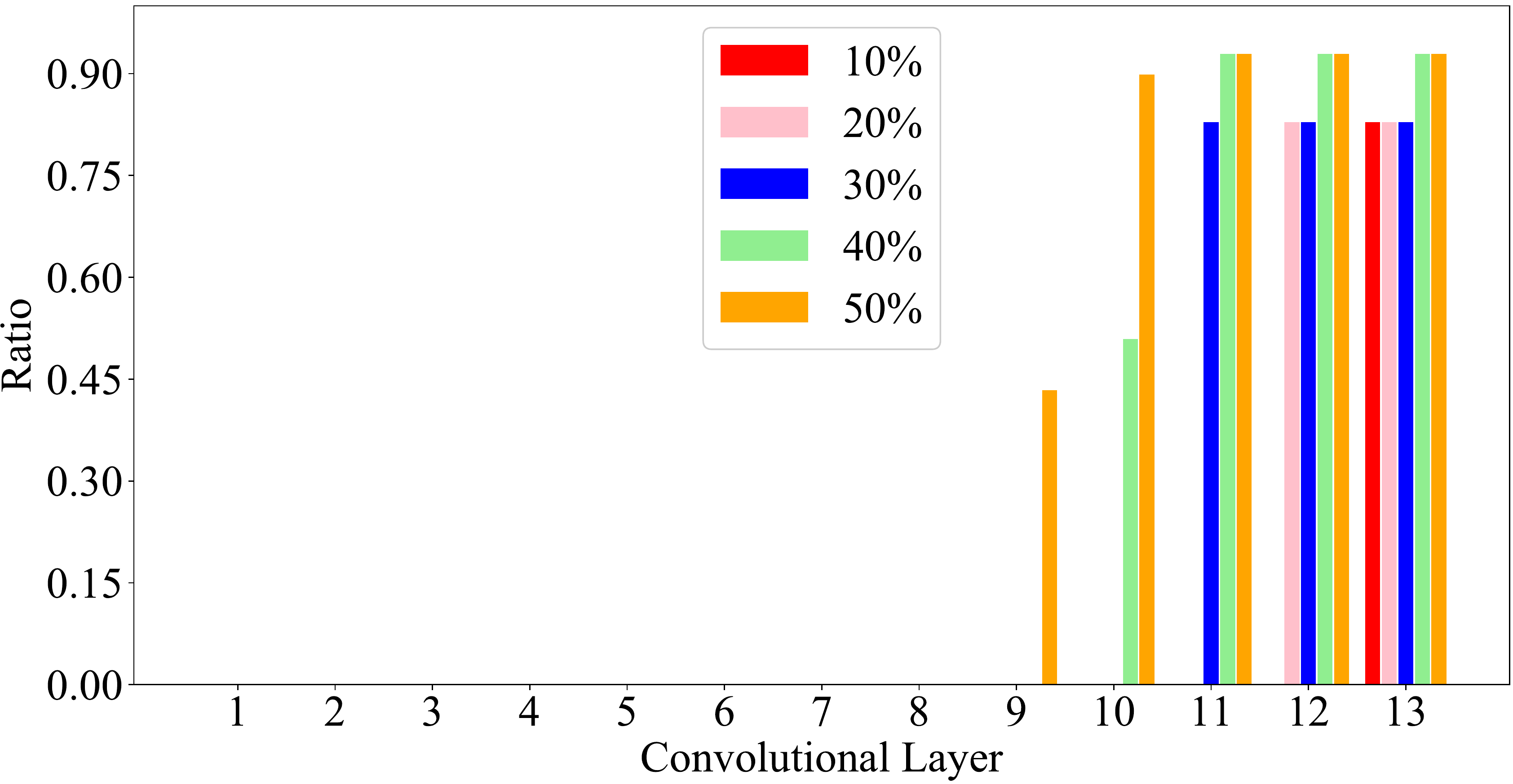}
        \end{minipage}
          }
      \subfigure[]{
        \begin{minipage}[t]{0.48\linewidth}
          \centering
          \includegraphics[width=2.8in]{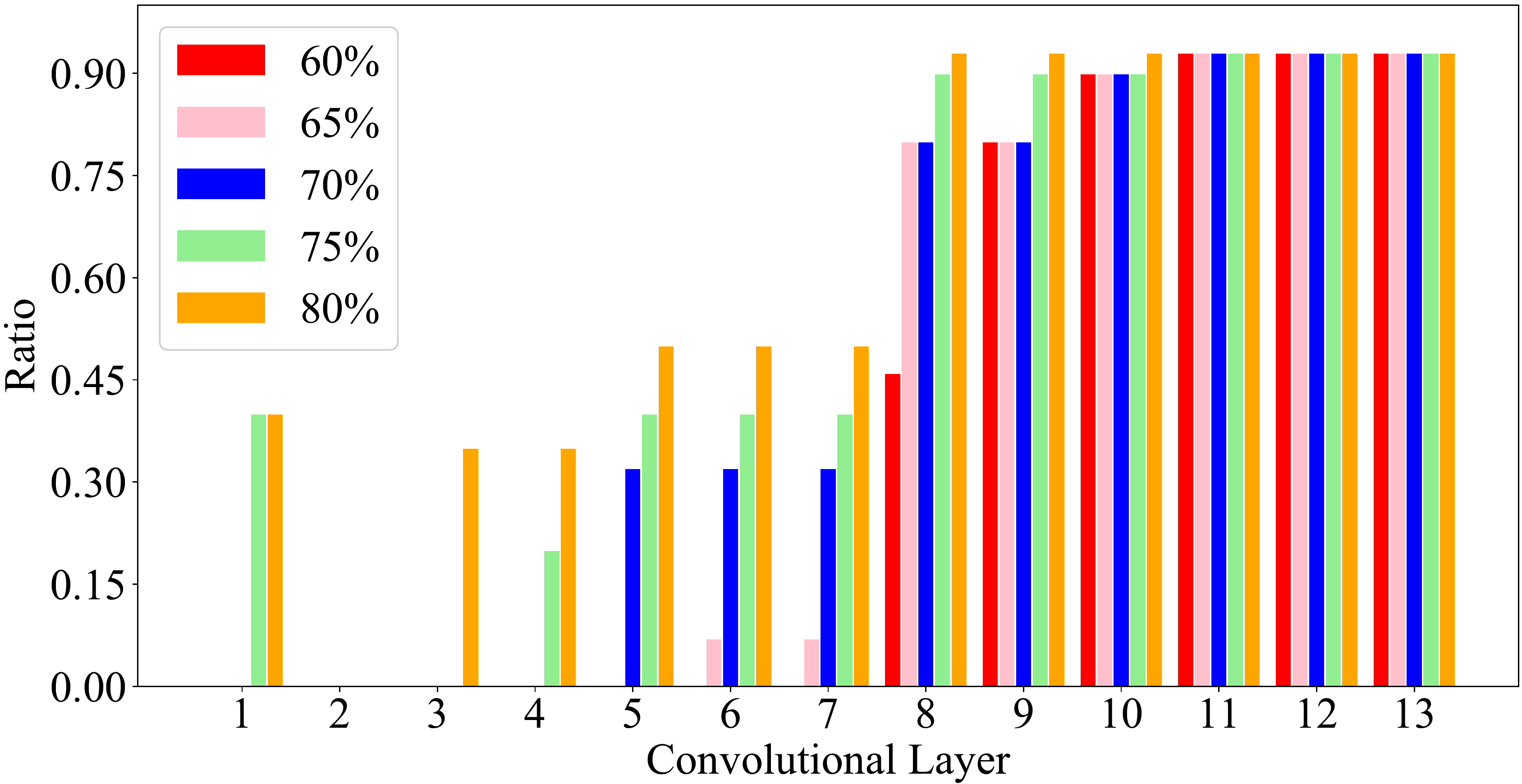}
        \end{minipage}%
      }%
      \caption{The detailed pruning ratio of each convolutional layer for VGG-16 on CIFAR-10. "1-13" in x-axis stands for the index of convolutional layers. In (a), we improve pruning ratio from 10\% to 50\%. In (b), the network is pruned from 60\% to 80\%. Apparently, insensitive layers are pruned heavily.}
      \label{tab:detailedvgg16}
    \end{figure*}

    CIFAR-10 dataset contains 10 categories of natural images, and 50000 and 10000 pictures are included in training and testing sets, respectively. VGG-16 is designed with 13 convolutional and 3 connection layers in feature extraction and classification modules, respectively. For saving computation resource, the classifier is decreased to $[512 \times 512 \times 10]$ for baseline model, and learning ratio is set to 0.01, 0.005 and 0.001 with every 50 epochs. Meanwhile, the BatchNorm layers are added after each convolutional layer. At first, we compared the results of hierarchy pruning when original models were trained with 50, 100 and 150 epochs respectively in Fig. \ref{tab:vgg16pruning}., and we observed that all layers showed similar tendency for different trained models during the hierarchy pruning process, which meant we could disentangle the importance of each layer even when the model was trained imperfectly. We calculated the score of sensitiveness for each convolutional layer on Table \ref{score_of_different_models}. Clearly, the score of each layer among all three different trained models were different, but the relative difference across all layers could keep stable. For example, the score of Conv2 was $2.0\times 10^{-3}$, $7.7\times 10^{-3}$, and $6.9\times 10^{-3}$ for the models that were trained with 50, 100 and 150, respectively, however, Conv2 was easy to be specified with the highest importance because its score was large than other layers obviously. Therefore, the consistency of importance evaluation for the each layer was be verified. Then we would explore the efficiency of SBPF on the model with 150 training epochs.
    Fig. \ref{tab:vgg16pruning}. (c) and (f) presented the untrained and retrained test accuracy of hierarchy pruning when original VGG-16 was trained fully. Notably, the accuracy restoration will decrease with the increasing of pruning ratio for each layer, and specific calculation of sensitiveness was listed Table \ref{tab:vgg16score} (a). The Conv2 gets the highest score, which is the most sensitive layer. "Conv1, Conv3, Conv4" and Conv5, Conv6, Conv7 Conv8" could be assembled to two groups respectively according to Equation (\ref{layers_grouping}). At last, the remaining layers were grouped together because they showed smaller score of sensitiveness. The grouped layers were depicted as:
     \begin{itemize}
        \item \textbf{$G_{vgg16}$:}\\
        \fbox {\shortstack[l]{Conv2\\
        Conv1, Conv3, Conv4\\
        Conv5, Conv6, Conv7, Conv8\\
        Conv9, Conv10, Conv11, Conv12, Conv13}}
    \end{itemize}
    Then we pruned the model by iterative strategy in Algorithm \ref{iterative-pruning}. For retraining the slimmed model efficiently, the learning ratio should be revised to 0.005 at beginning 50 epochs and 0.001 for following epochs. In Fig. \ref{tab:vgg16retraining}., we found that the pruned model may perform a better result than baseline when pruning ratio was less than 60\%. Even we increased the pruning ratio to 70\%, the pruned model could perform well without accuracy sacrificing. However, the representation capability of pruned model fell rapidly when we continued to increase the pruning ratio. When the pruning ratio attained to 80\%, there could be 4\% loss in accuracy. The pruned results were listed in Table \ref{benchmarks_pruning}. Additionally, Fig. \ref{tab:detailedvgg16}. showed the details about final iterative pruning process. At beginning, layers in insensitive group were pruned, and others kept unchanged. With the increasing of pruning ratio, remaining layers would be
    pruned by the degree of sensitiveness iteratively.
\begin{table*}[htbp]
\tiny
  \centering
  \caption{The sensitiveness of each convolutional layer for different pre-trained VGG-16, Conv-4 and ResNet-18.}
     \setlength{\tabcolsep}{1.8mm}{
        \begin{tabular}{|p{0.8cm}|p{1.2cm}|p{0.1cm}|p{0.7cm}|p{0.7cm}|p{0.7cm}|p{0.7cm}|p{0.7cm}|p{0.7cm}|p{0.7cm}|p{0.7cm}|p{0.7cm}|p{0.7cm}|p{0.7cm}|p{0.7cm}|p{0.7cm}|}
    \hline
    Model & Training Epochs & $\gamma$ & \multicolumn{13}{c}{$\emph{S}$} \\
    \hline
          &       &       & \multicolumn{1}{l}{Conv1} & \multicolumn{1}{l}{Conv2} & \multicolumn{1}{l}{Conv3} & \multicolumn{1}{l}{Conv4} & \multicolumn{1}{l}{Conv5} & \multicolumn{1}{l}{Conv6} & \multicolumn{1}{l}{Conv7} & \multicolumn{1}{l}{Conv8} & \multicolumn{1}{l}{Conv9} & \multicolumn{1}{l}{Conv10} & \multicolumn{1}{l}{Conv11} & \multicolumn{1}{l}{Conv12} & \multicolumn{1}{l}{Conv13} \\
          \hline
    \multirow{3}[0]{*}{VGG-16} & \centering 50    & \multirow{9}[0]{*}{\centering $\frac{2}{3}$}  & 2.2E-04     &   2.0E-03     &   4.3E-04     &   3.5E-04     &    -1.1E-03    &  -1.4E-03      &   -1.6E-03     &   -1.8E-03     &   -2.8E-03     &  -3.0E-03      &   -2.8E-03     &      -3.4E-04 & -4.0E-04  \\
    \cline{2-2}
          & \centering 100   &  &4.8E-03     &  7.7E-03     &    5.3E-03   & 5.1E-03      &3.7E-03       &  3.3E-03     &    3.1E-03   & 2.7E-03      &  2.0E-03     &2.0E-03       &1.5E-03       & 7.6E-04      &  4.6E-05  \\
          \cline{2-2}
          & \centering 150   &       &    4.7E-03   &   6.9E-03    &    5.2E-03   &5.2E-03       & 3.8E-03      & 3.7E-03      &3.6E-03       & 3.2E-03      & 2.4E-03      &    2.5E-03   &    1.2E-03   &    8.5E-04   &1.3E-05  \\
          \cline{1-2}
    \multirow{3}[0]{*}{Conv-4} &\centering 5     &       &  8.9E-03     &  2.4E-04     &  -1.6E-04      & -2.1E-04       &     &       &       &       &       &       &       &       &  \\
    \cline{2-2}
          & \centering 10    &       &   1.0E-02    &  1.1E-03     &-1.0E-03       &-9.2E-04       &       &       &       &       &       &       &       &       &  \\
          \cline{2-2}
          & \centering 20    &       &1.0E-02       &1.5E-03       &-2.1E-04       & -4.5E-04      &       &       &       &       &       &       &       &       &  \\
          \cline{1-2}
    \multirow{3}[0]{*}{ResNet-18} & \centering 50    &       & 3.2E-04      &    2.9E-04   &    2.3E-03   &     4.4E-04  &   3.0E-04    &  -1.4E-03     &    -1.7E-03   &-1.6E-03       &       &       &       &       &  \\
    \cline{2-2}
          & \centering 100   &       & 7.2E-04      &    4.6E-04   &    2.5E-03   &     6.6E-04  &   5.7E-04    &  -1.0E-03     &    -1.3E-03   &-1.1E-03&       &       &       &       &  \\
          \cline{2-2}
          & \centering 150   &      & 2.9E-03      &    2.4E-03   &    4.4E-03   &     3.0E-03  &   3.1E-03    &  6.2E-04     &    1.3E-03   &1.0E-03&       &       &       &       &  \\
          \hline
    \end{tabular}}%
  \label{score_of_different_models}%
\end{table*}%
  \subsection{Customized Conv-4 on MNIST}
\begin{figure}[htbp]
      \centering
      \subfigure[Untrained model]{
        \begin{minipage}[t]{1\linewidth}
          \centering
          \includegraphics[width=2.0in]{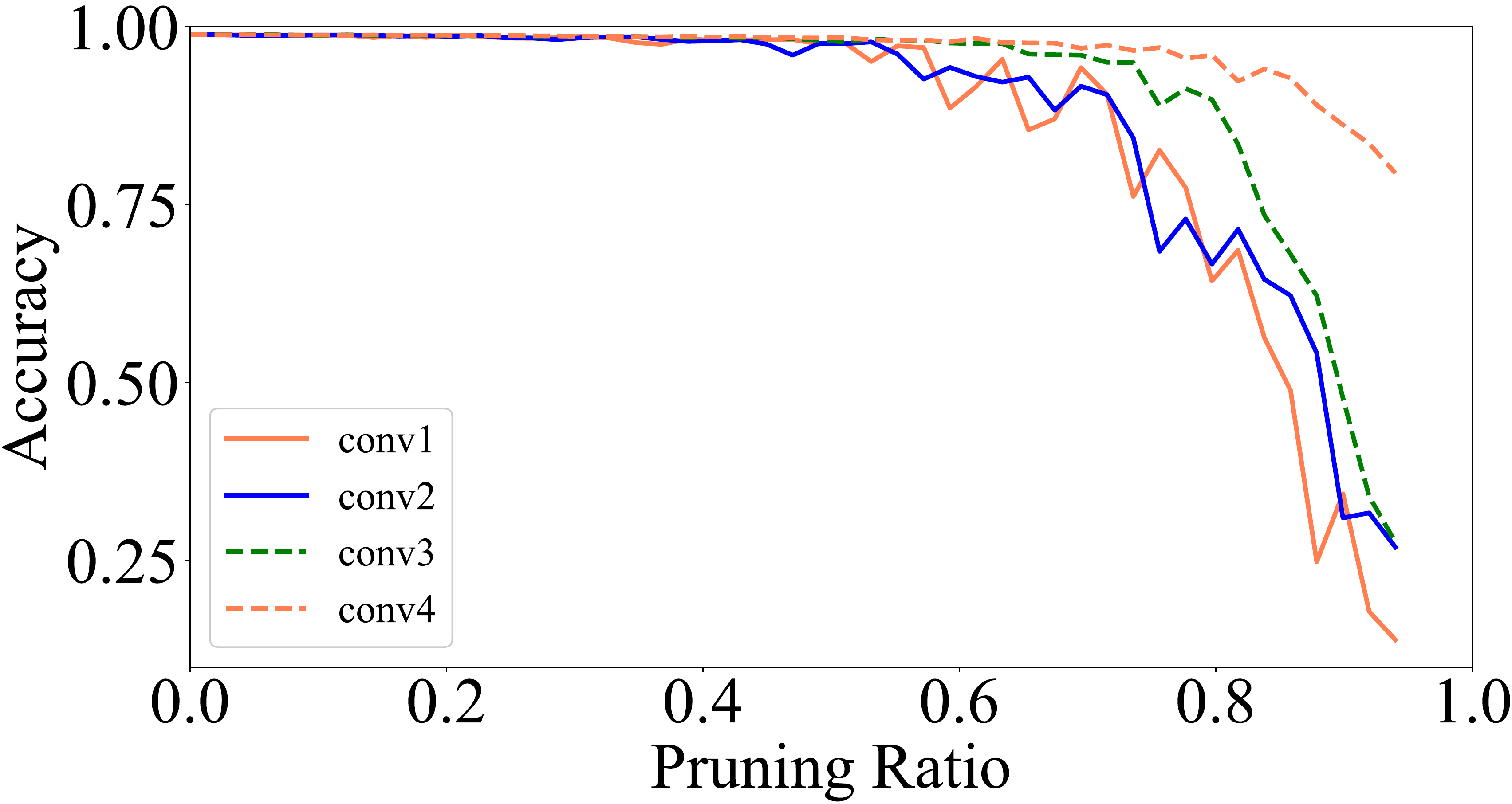}
        \end{minipage}
      }
      \subfigure[Retrained model]{
        \begin{minipage}[t]{1\linewidth}
          \centering
          \includegraphics[width=2.0in]{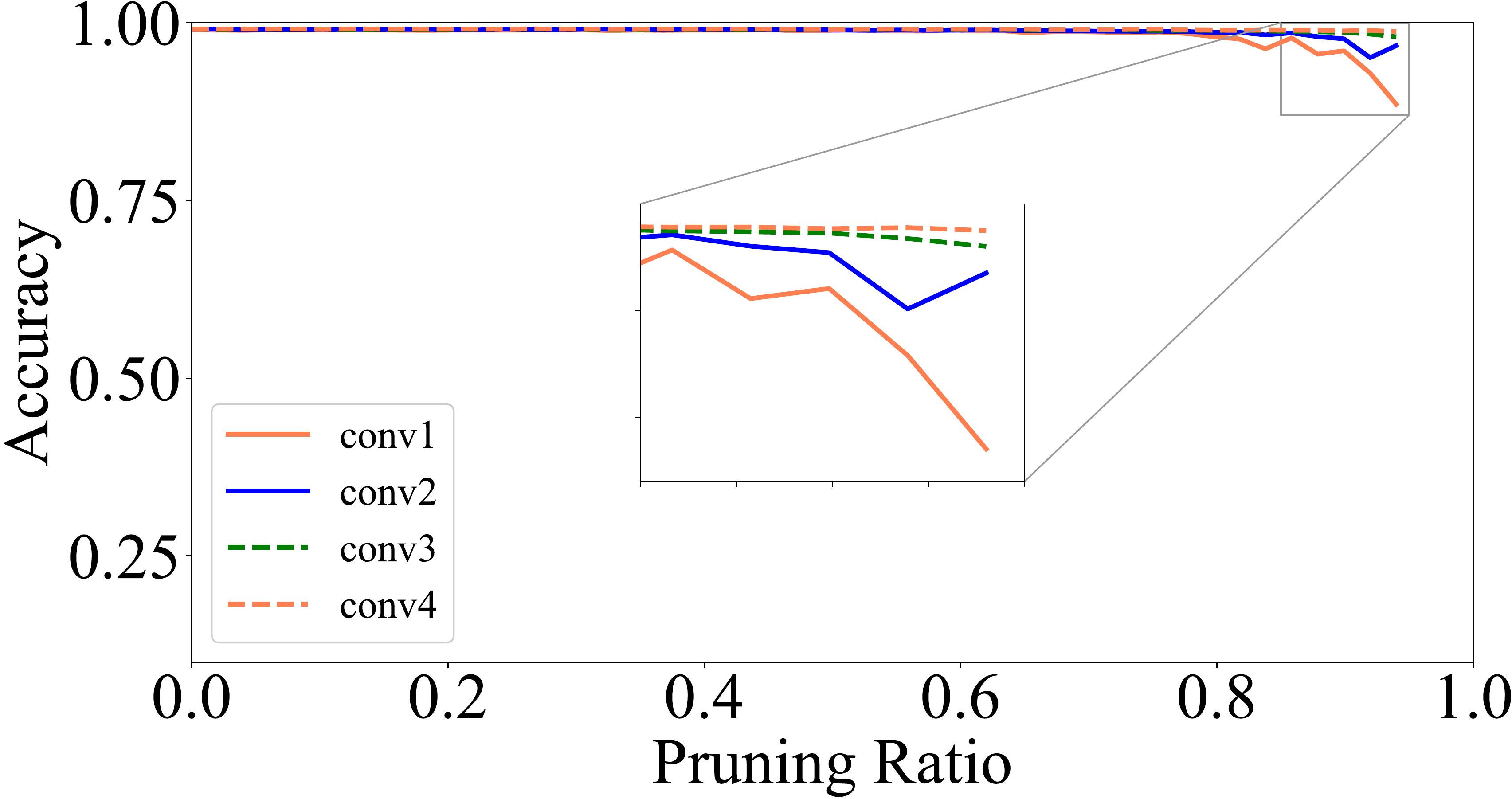}
        \end{minipage}
      }
      \centering
      \caption{Hierarchy pruning for Conv-4 on MNIST. The (a) shows untrained test accuracy and the (b) presents retrained test accuracy.}
      \label{tab:Conv4pruning}
    \end{figure}

    MNIST dataset includes 10 categories of handwritten numbers with smaller scale, which could be trained and tested with a simple network. Thus, we shrinked convolutional layers of VGG-16 to 4 (Conv-4) as $[64\times128\times256\times512]$ , and maxpool was reduced to 2 relatively. The size of classifier was maintained same as above VGG-16. At first, the sensitiveness of each convolutional layer was tested on Table \ref{score_of_different_models} when Conv-4 was trained variously. When the model was trained with 10 and 20 epochs, the score of sensitiveness for each layer could keep similar. when the model was trained with 5 epochs, the score of each layer was quite different with the fully training model, but the relative difference across all layers can stay similar as the fully trained model. For example, Conv1 showed the highest score for all three different trained models, and other layers all showed relative lower sensitiveness. Then we tested our algorithm on the fully trained model. Notably, the learning ratio was set as 0.01 across total 20 epochs, and the hierarchy pruning process was depicted as Fig. \ref{tab:Conv4pruning}., from which we could see that the fluctuation was smaller than VGG-16. The specific score of fully trained model was illustrated on Table \ref{tab:vgg16score} (b). The score of "Conv1" was large than others apparently, then we
     \begin{figure*}[htbp]
      \centering
      \subfigure[]{
        \begin{minipage}[t]{0.31\linewidth}
          \centering
          \includegraphics[width=2.0in]{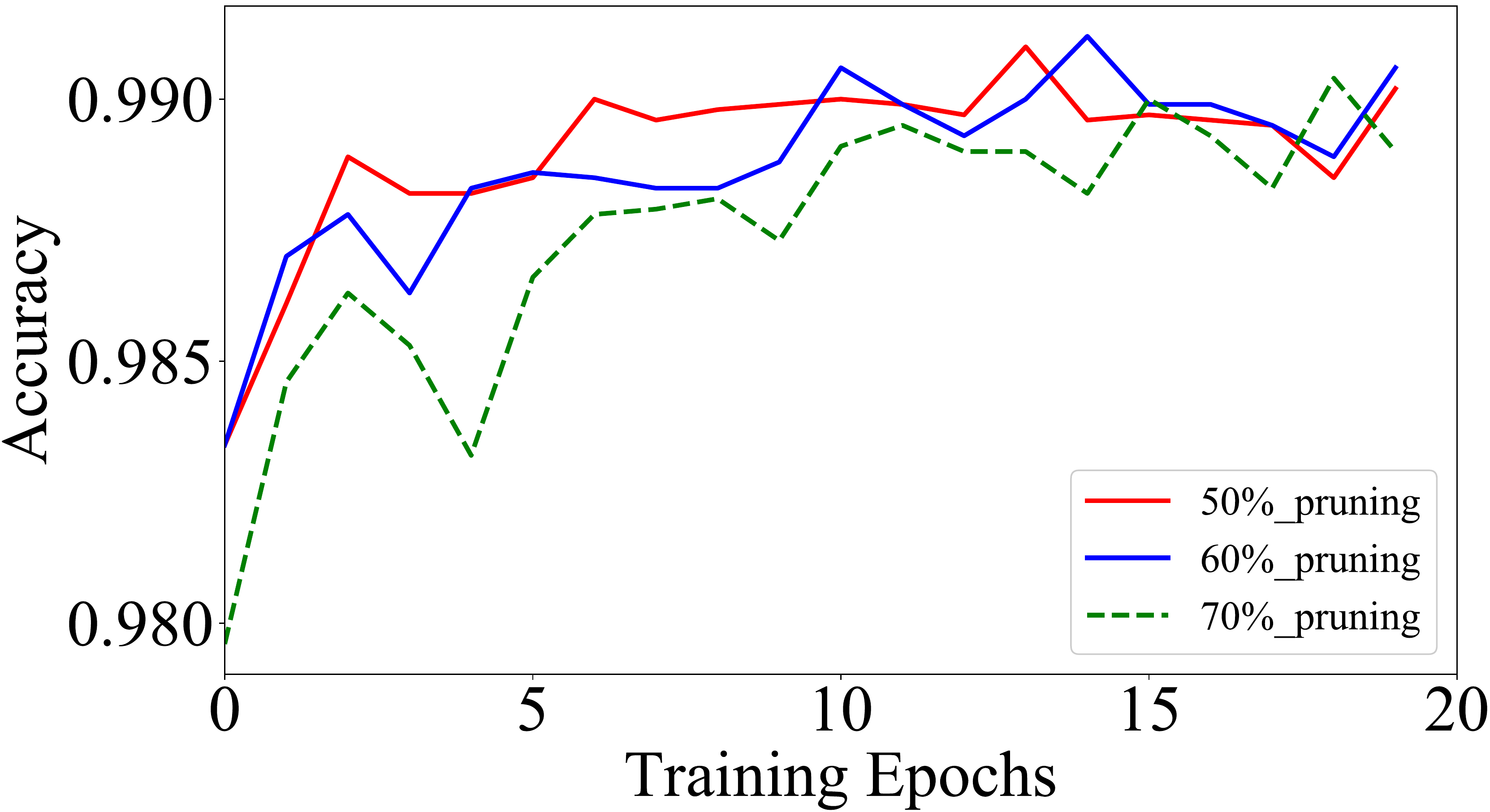}
      \end{minipage}
      }
      \subfigure[]{
        \begin{minipage}[t]{0.31\linewidth}
          \centering
          \includegraphics[width=2.0in]{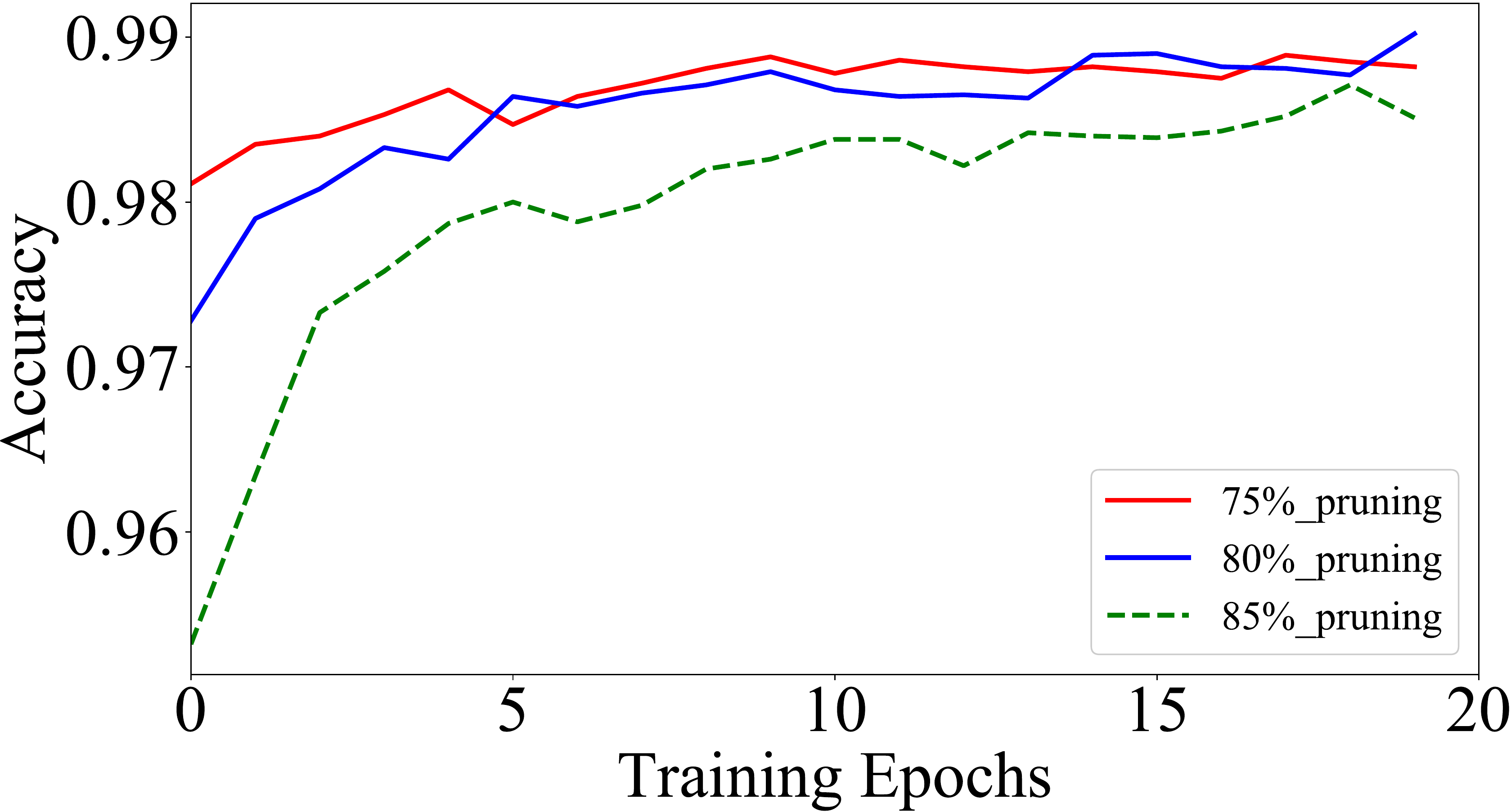}
        \end{minipage}
      }
      \subfigure[]{
        \begin{minipage}[t]{0.31\linewidth}
          \centering
          \includegraphics[width=2.0in]{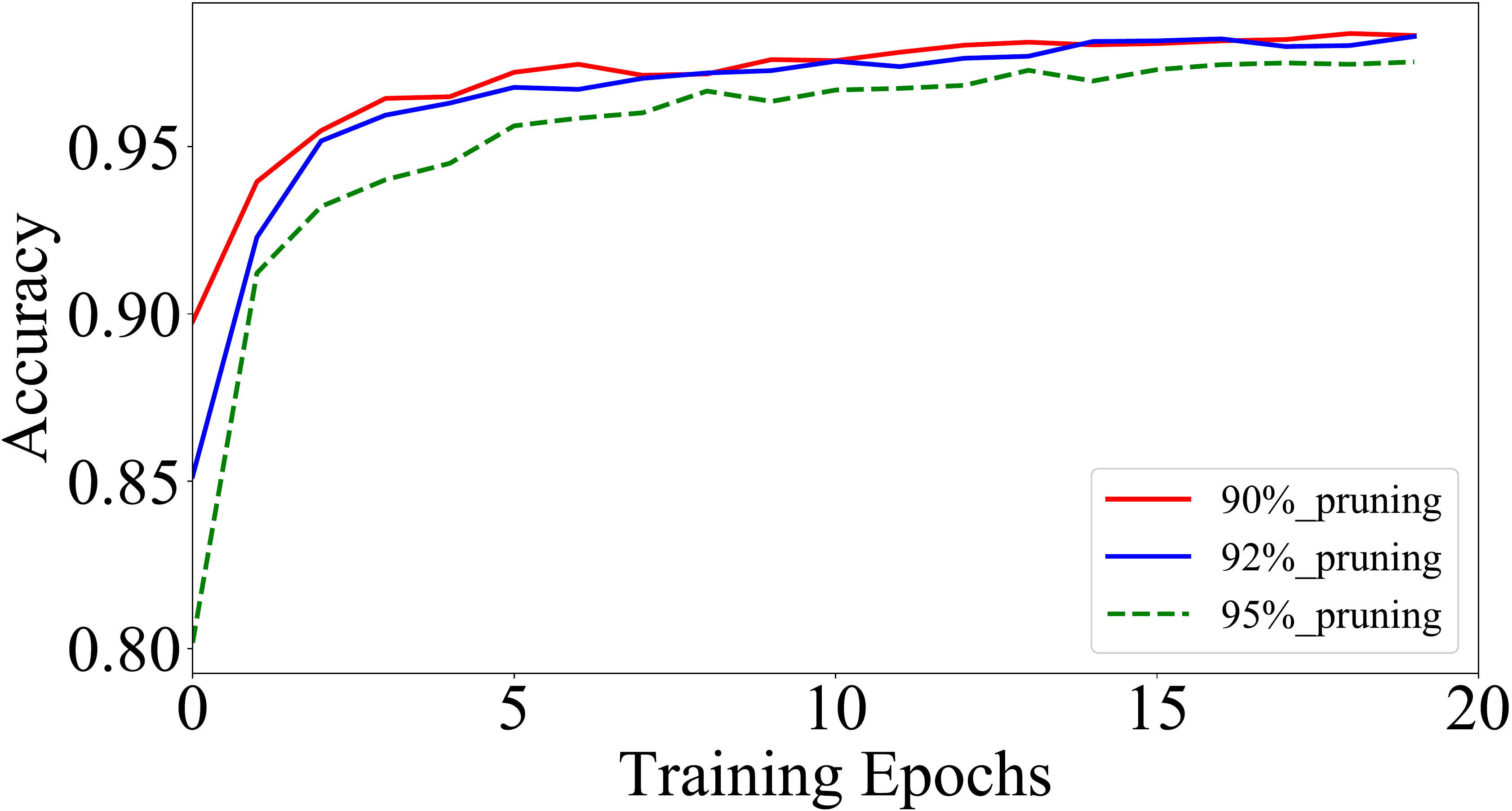}
        \end{minipage}
      }
      \caption{Retrained results for Conv-4 on MNIST with iterative pruning. In (a), the pruning ratio is 50\%, 60\% and 70\%. Then the pruning ratio increases to 75\%, 80\%, 85\% in (b) and 90\%, 92\%, 95\% in (c), respectively. Pruning ratio is larger than VGG-16 with CIFAR-10.}
          \label{tab:convretraining}
    \end{figure*}grouped "Conv2, Conv3, Conv4" together, and separated "Conv1" independently. The grouping was showed as:
       \begin{itemize}
        \item \textbf{$G_{Conv-4}$:}\\
        \fbox {\shortstack[l]{Conv1\\
        Conv2, Conv3, Conv4}}
    \end{itemize}
    Subsequently, iterative pruning for the original model with fully training was explored as Fig. \ref{tab:convretraining}., and the pruned model with 85\% pruning ratio could recover baseline accuracy after carefully retraining. However, the accuracy dropped quickly if more filters were removed, and 2\% loss in accuracy could appear for the pruned model when we improved pruning ratio to 95\%. Finally, the overall compression statistics were listed on Table \ref{benchmarks_pruning}, which showed that there were more redundant filters in customized Conv-4 than VGG-16.
  \subsection{ResNet-18 on CIFAR-100}
\begin{figure}[htbp]
      \centering
      \subfigure[Untrained model]{
        \begin{minipage}[t]{1\linewidth}
          \centering
          \includegraphics[width=2.0in]{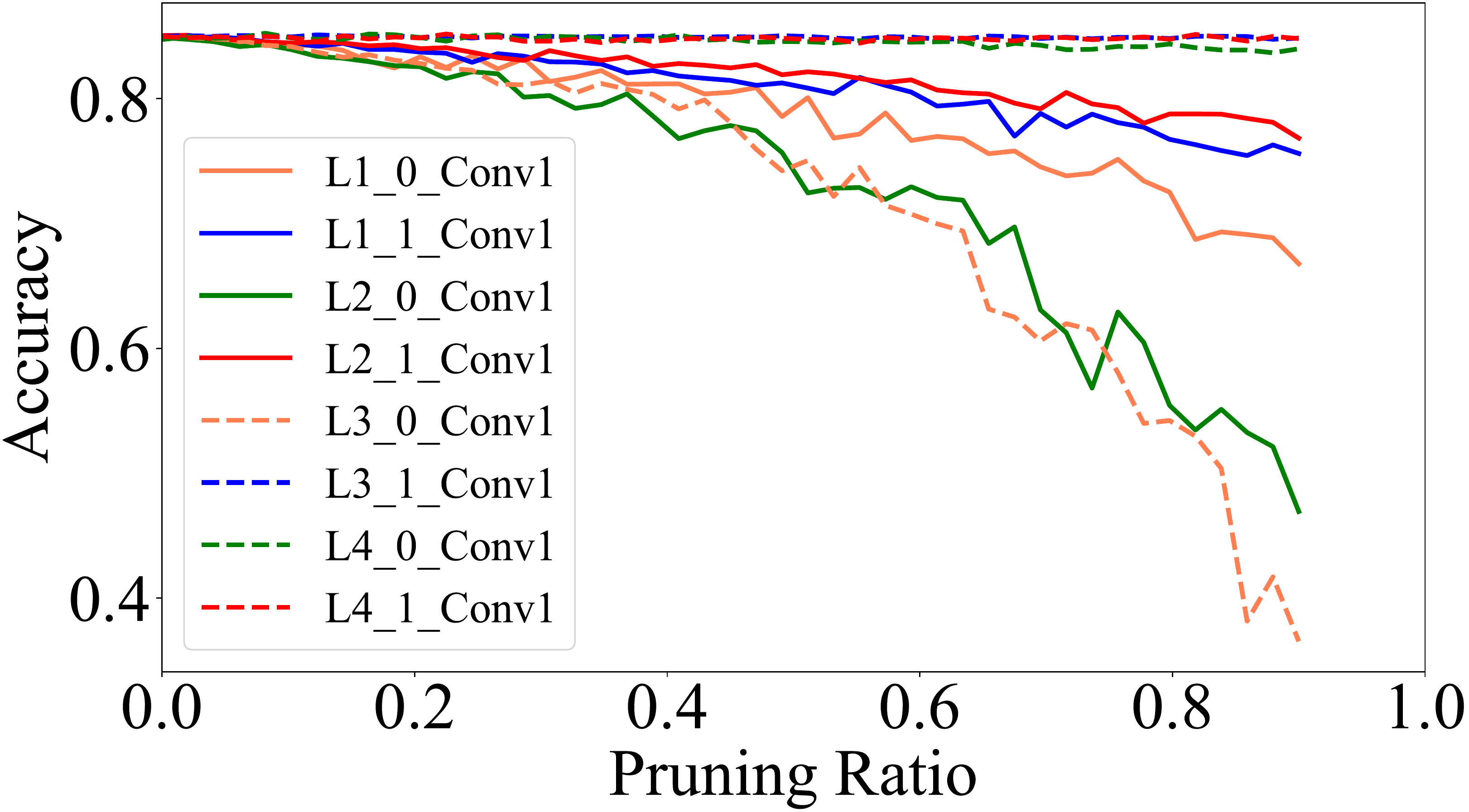}
        \end{minipage}
      }
      \subfigure[Retrained model]{
        \begin{minipage}[t]{1\linewidth}
          \centering
          \includegraphics[width=2.0in]{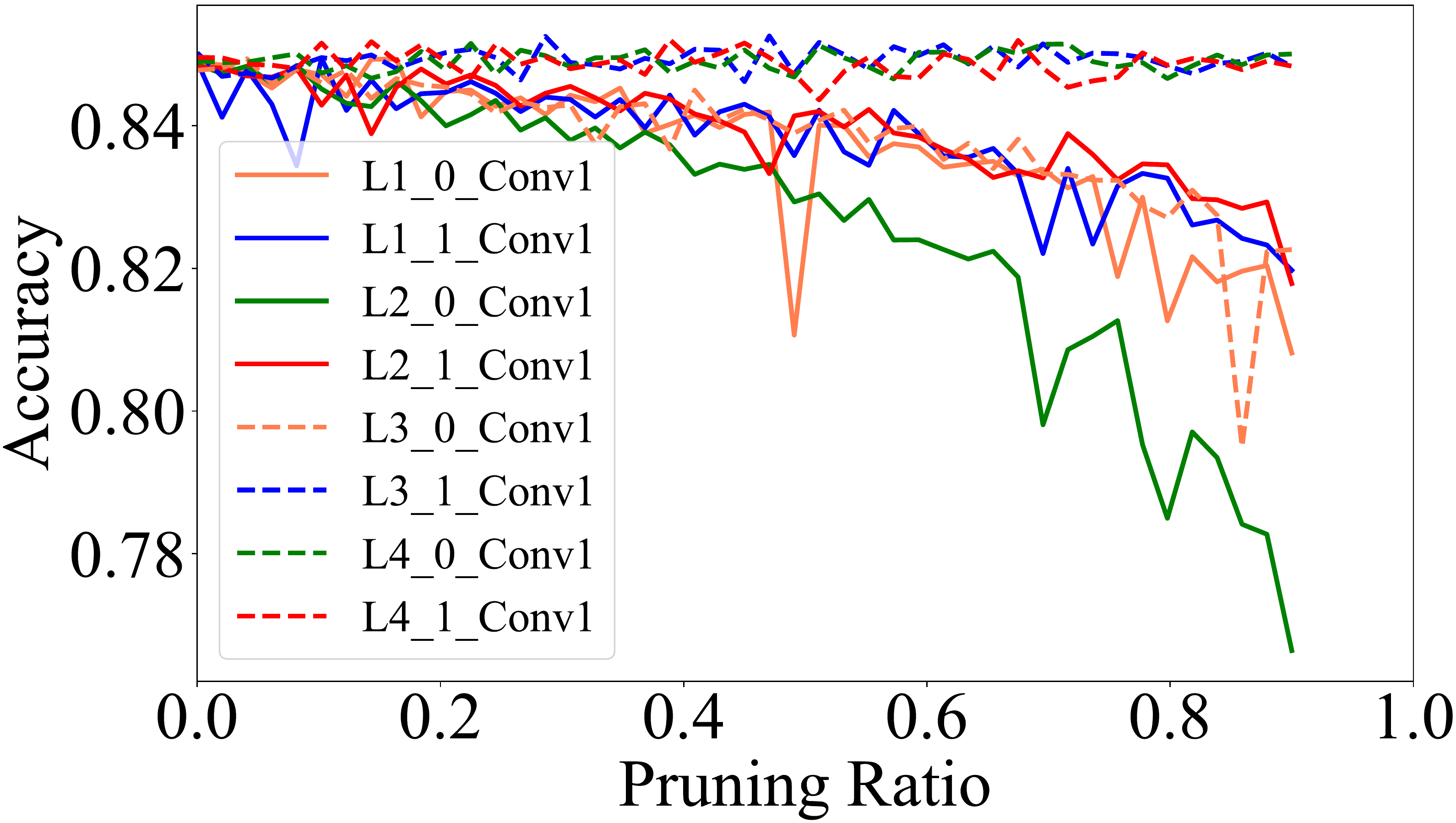}
        \end{minipage}
      }
      \centering
      \caption{Hierarchy pruning for ResNet-18 on CIFAR-100. The (a) shows untrained test accuracy and the (b) presents retrained test accuracy.}
      \label{tab:resnetpruning}
    \end{figure}

\begin{figure}[htbp]
      \centering
      \subfigure{
        \begin{minipage}[t]{1\linewidth}
          \centering
          \includegraphics[width=2.0in]{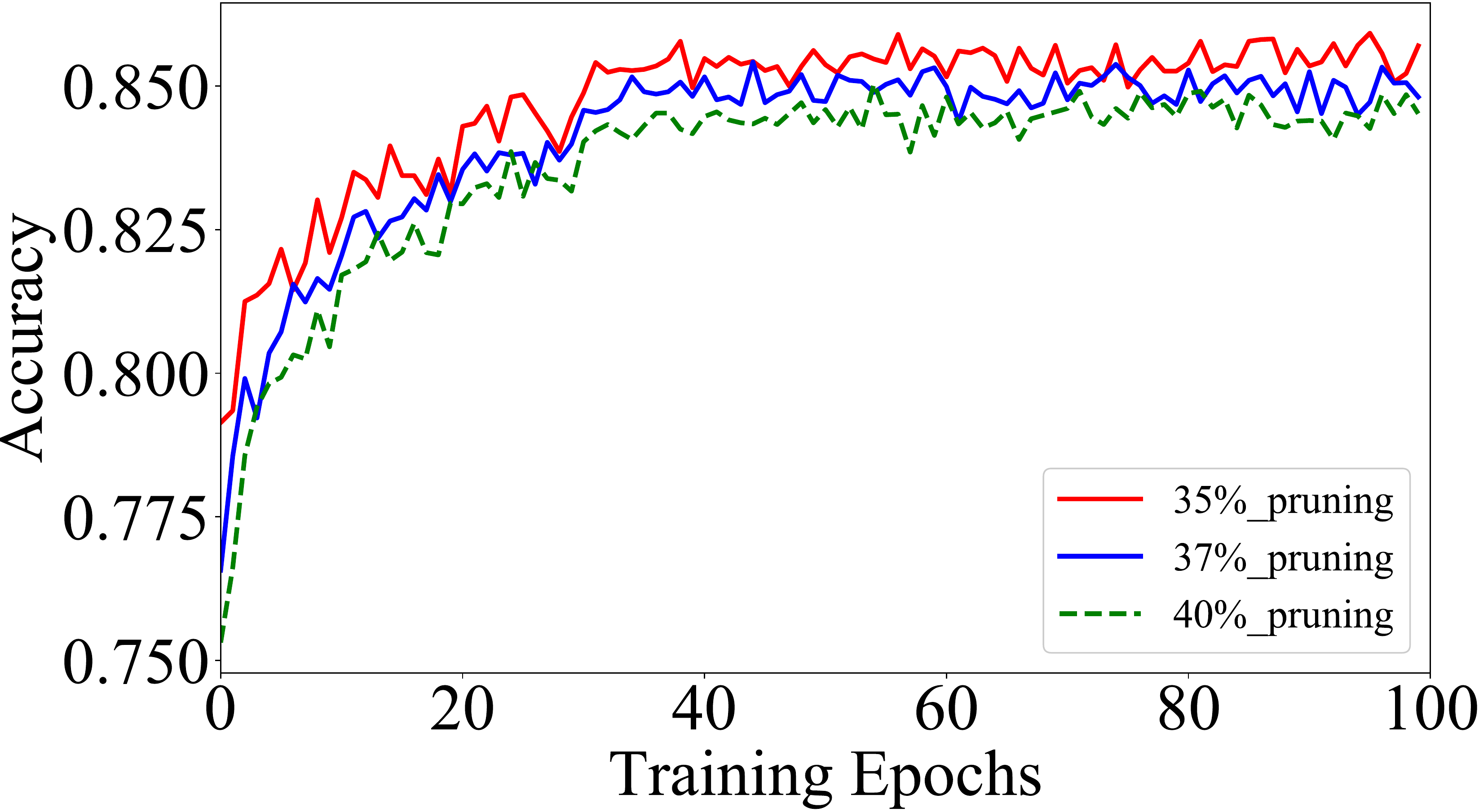}
        \end{minipage}
      }
      \caption{The retrained results with 35\%, 37\% and 40\% pruning ratio for ResNet-18 on CIFAR-100 dataset.}
      \label{tab:resnet18retraining}
    \end{figure}

    VGG-16 and Conv-4 are both chain structures without skip connection in feature extraction block. Therefore, we extended the pruning framework on ResNet-18 over CIFAR-100, which was covered by 8 residual blocks with 16 convolutional layers. Similarly, the images in CIFAR-100 have same size as CIFAR-10 while only extending categories to 100. Notably, overall pruning ratio on ResNet-18 would be lower than VGG-16 and Conv-4 because the skip connection in each block imposed strictly dimension restriction to prune whole network, and only the first convolutional layer could be slimmed. For simplifying explanation, we renamed the pruned layers as Conv1 to Conv8, and the learning ratio of the baseline model was set to 0.01, 0.005, 0.0001 for every 50 epochs, respectively. At first, we applied hierarchy pruning on different pre-trained models. The explored sensitiveness is displayed on Table \ref{score_of_different_models}. Same as VGG-16 and Conv-4, the sensitiveness of each layer across the different trained models kept similar tendency, which illustrated that our pruning framework could maintain the consistency of importance evaluate for each layer without considering the parameters of individual filter. Then we plotted the results of hierarchy pruning of ResNet-18 when the original model was fully trained in Fig. \ref{tab:resnetpruning}., the deeper layers were less important because the accuracy of corresponding pruned models fluctuated slightly across the whole hierarchy pruning process. Then we calculated the specific score of reliability and stability on Table \ref{tab:vgg16score} (c). Correspondingly, "Conv1, Conv2, Conv4, Conv5", "Conv6, Conv7, Conv8" could be classified to two groups due to similar scores while Conv3 could be grouped individually due to the highest score. Then the specific layers clustering was concluded as:
       \begin{itemize}
        \item \textbf{$G_{resnet18}$:}\\
        \fbox {\shortstack[l]{Conv3\\
        Conv1, Conv2, Conv4, Conv5\\
        Conv6, Conv7, Conv8}}
    \end{itemize}
    In final iterative pruning stage, the learning ratio was set to 0.0005 at beginning 50 epochs, and following 50 epochs would be set to 0.0001. Fig. \ref{tab:resnet18retraining}. showed the iterative process. When we set the pruning ratio as 35\%, the results of pruned model may outperform baseline models, and the retrained pruned network can reach to baseline accuracy even with 37\% filters removed. But when we increased the pruning ratio a bit to 40\%, there could be obvious declining in accuracy during whole pruning process. The conclusion about the pruned outcomes were listed on Table \ref{benchmarks_pruning} (c), which illustrated that our pruning framework achieved impressive pruning results even with dimension restriction.
     \begin{figure}[htb]
      \centering
      \subfigure[$\mathscr{C}_1$-norm VS randomness VS $\mathscr{C}_2$-norm]{
        \begin{minipage}[t]{1\linewidth}
          \centering
          \includegraphics[width=2.0in]{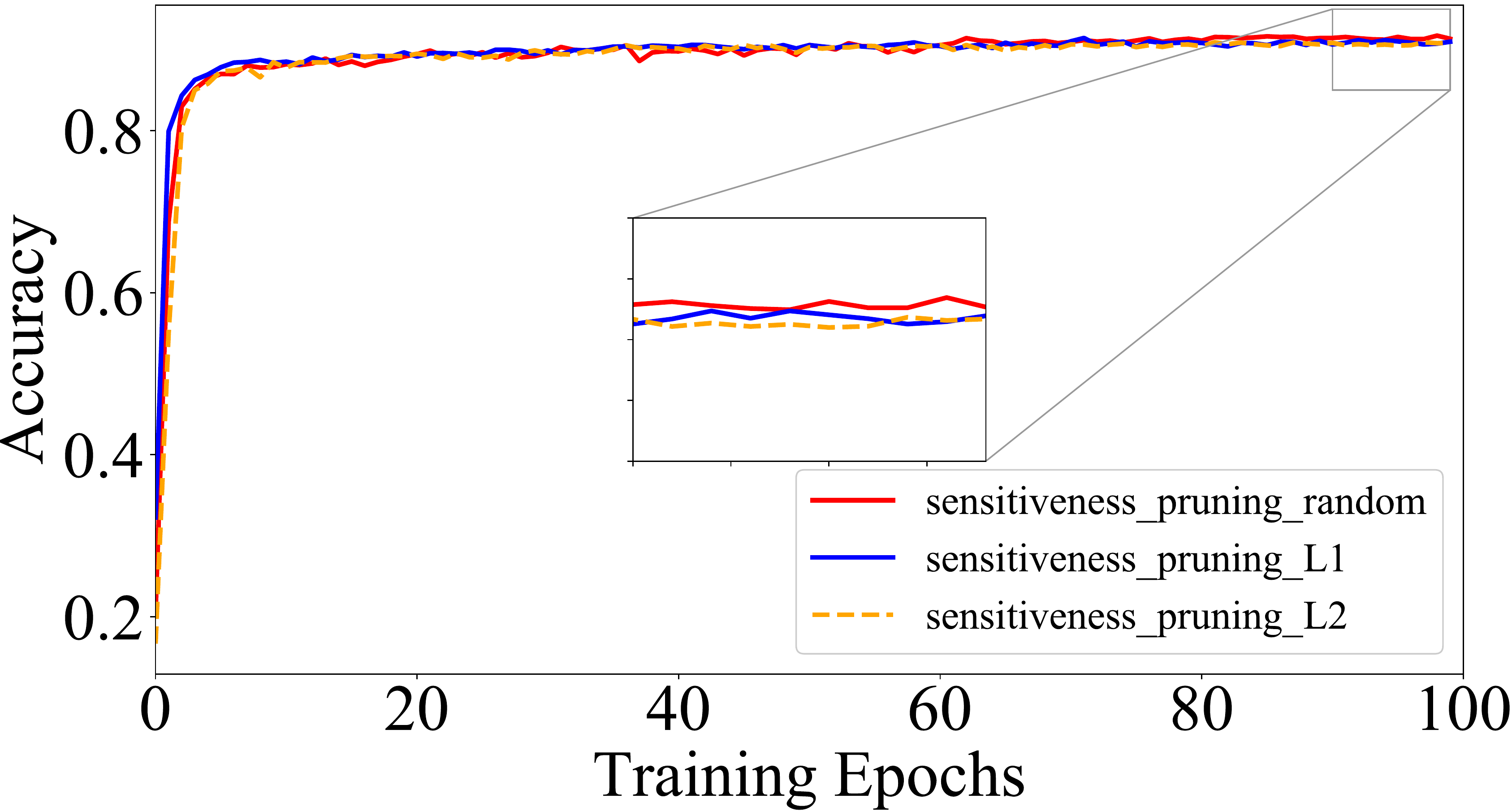}
        \end{minipage}
      }
            \caption{The comparison of different pruning strategies for VGG-16 on CIFAR-10.}
      \label{tab:comparision}
    \end{figure}
        \begin{figure}[htb]
        \centering
        \subfigure{
          \begin{minipage}[t]{1\linewidth}
            \centering
          \includegraphics[width=2.0in]{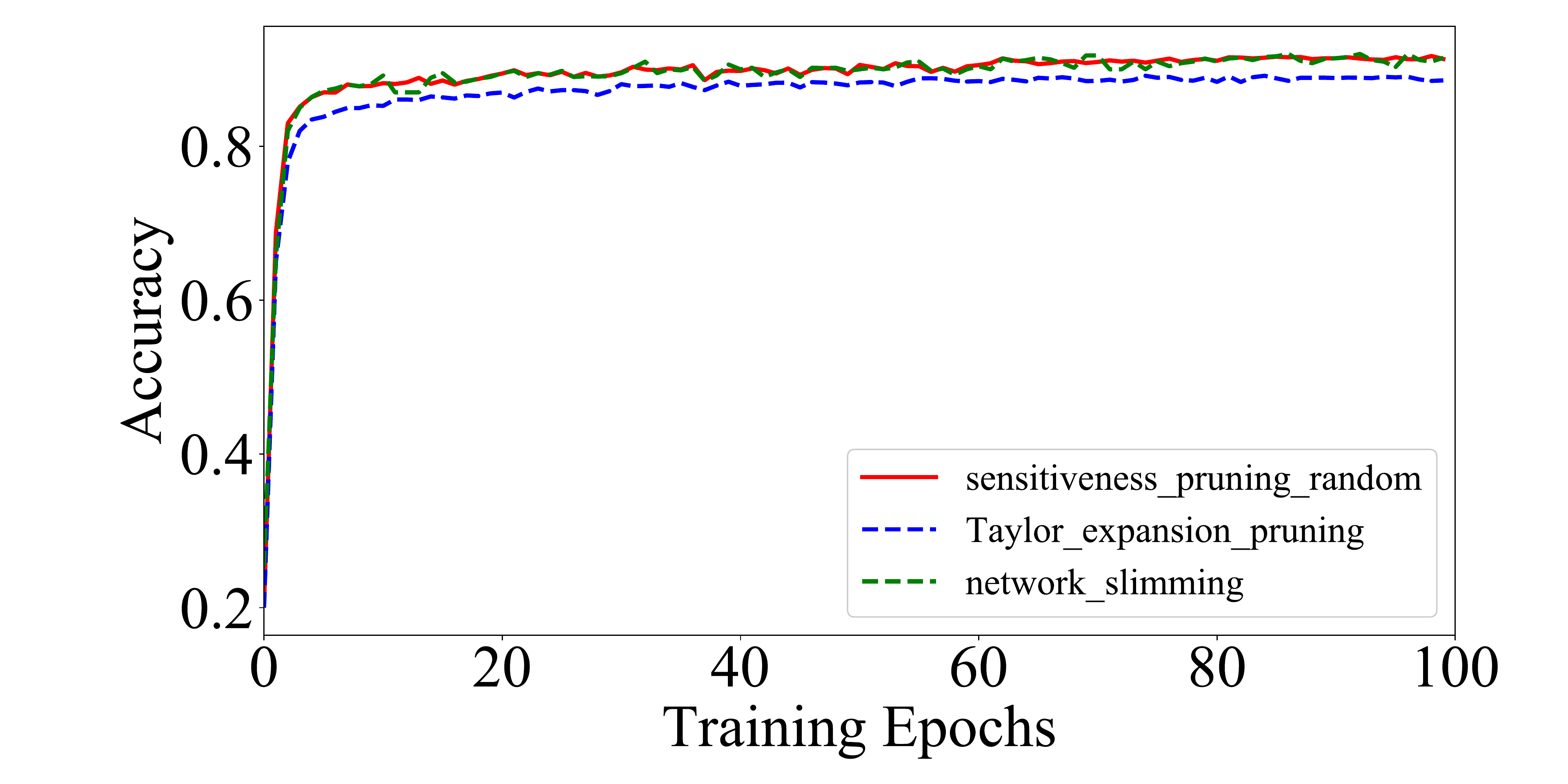}
          \end{minipage}
        }
        \caption{The retrained results based on Taylor expansion, network slimming and sensitiveness pruning criteria for VGG-16 on CIFAR-10 with 65\% pruning ratio.}
        \label{tab:taylor_sensitiveness}
      \end{figure}
\renewcommand\arraystretch{1.0}
      \begin{table*}[htb]
    \scriptsize
      \centering
      \caption{The distribution of filters when the pruning ratio is set to 65\%. T1-T5 means the pruning iterations for Taylor expansion method, and T collects the total number of pruned filters for after final pruning. N-S Stands For Network Slimming Method, And $\emph{S}$ Stands For our method.}
       \setlength{\tabcolsep}{2.5mm}{
        \begin{tabular}{p{1.3cm}p{0.5cm}p{0.5cm}p{0.5cm}p{0.5cm}p{0.5cm}p{0.5cm}p{0.5cm}p{0.5cm}p{0.5cm}p{0.5cm}p{0.5cm}p{0.5cm}p{0.5cm}p{0.5cm}r}
        \toprule
        Criteria & \multicolumn{13}{c}{Filters Pruned}                                                                   & Total \\
        \midrule
                & Conv1 & Conv2 & Conv3 & Conv4 & Conv5 & Conv6 & Conv7 & Conv8 & Conv9 & Conv10 & Conv11 & Conv12 & Conv13 &       &  \\
        \midrule
        T1 (13\%) & 0     & 1     & 4     & 2     & 9     & 14    & 7     & 46    & 40    & 53    & 76    & 82    & 215   &       &  \\
        T2 (26\%) & 7     & 4     & 5     & 6     & 8     & 12    & 13    & 53    & 61    & 62    & 114   & 92    & 112   &       &  \\
        T3 (39\%) & 3     & 10    & 8     & 11    & 34    & 18    & 30    & 70    & 85    & 77    & 93    & 68    & 42    &       &  \\
        T4 (52\%) & 6     & 2     & 15    & 14    & 34    & 33    & 36    & 82    & 80    & 74    & 54    & 75    & 44    &       &  \\
        T5 (65\%) & 5     & 12    & 15    & 19    & 37    & 47    & 41    & 72    & 82    & 73    & 50    & 66    & 30    &       &  \\
        \midrule
        T (65\%)& 21    & 29    & 47    & 52    & 122   & 124   & 127   & 323   & 348   & 339   & 387   & 383   & 443   & 2745  &  \\
        \midrule
		 N-S (65\%)& 12    & 0    & 0    & 0    & 0   & 0   & 6   & 307   & 474   & 503   & 510   & 508   & 427   & 2746  &  \\
		\midrule
        \textbf{\emph{S} (65\%)} & 0     & 0     & 0     & 0     & 0     & 18    & 18    & 410   & 410   & 461   & 476   & 476   & 476   & 2745  &  \\
        \bottomrule
        \end{tabular}
         }%
      \label{tab:tablecompare}
\end{table*}
\renewcommand\arraystretch{1.0}
 \begin{table*}[htb]
  \centering
  \scriptsize
  \caption{Results Of Different Pruning Methods For VGG-16 On CIFAR-10. The "Sensitiveness-$L_{1}$", "Sensitiveness-$L_{2}$" and "Sensitiveness-$L-{R}$" stands for sensitiveness based pruning framework with $L_{1}$, $L_{2}$ and random filters sorting strategies.}
      \centering
      \setlength{\tabcolsep}{4.0mm}{
    \begin{tabular}{|p{2.5cm}|p{2.2cm}|p{1.0cm}|p{1.5cm}|p{1.5cm}|p{1.0cm}|p{1.0cm}|}
    \hline
    \multirow{2}{2.5cm}{\centering Models(datasets)} & \multirow{2}{2.2cm}{\centering Methods} & \centering Filters (Pruned) & \centering Parameters (Pruned) & \centering FLOPs (MACs) (Pruned) &  \multirow{2}{1.0cm}{\centering Accuracy} & \multirow{2}{1.0cm}{\centering Constraints} \\
    \hline
    VGG-16(Baseline) &       &       & \centering 14.99M & \centering $3.1\times 10^8$ & \multirow{1}{1.0cm}{\centering 92.0\%}& \\
    \hline
    \multirow{12}[0]{*}{\centering VGG-16(CIFAR-10)} &\multirow{2}{2.2cm}{\centering Filters Pruning} & \multirow{12}{1.0cm}{\centering 65\%}  & \centering 4.5M (69.98\%) & \centering $1.9\times10^8$ (38.71\%) &  \multirow{2}{1.0cm}{\centering 92.2\%}&  \multirow{2}{1.0cm}{\centering -} \\
    \cline{2-2}\cline{4-7}
           & \multirow{2}{2.2cm}{\centering Taylor} &   & \centering 1.72M (88.52\%) & \centering $8\times 10^7$ (74.19\%) &  \multirow{2}{1.0cm}{\centering 90.3\%}& \multirow{2}{1.0cm}{\centering -} \\
          \cline{2-2}\cline{4-7}
          & \multirow{2}{2.2cm}{\centering Network Slimming} & &\centering 2.1M (85.98\%) &\centering $1.8\times 10^8$ (41.94\%) & \multirow{2}{1.0cm}{\centering 92.0\%}& \multirow{2}{1.0cm}{\centering Yes} \\
          \cline{2-2}\cline{4-7}
          & \multirow{2}{2.2cm}{\centering \textbf{Sensitiveness-$\mathscr{C}_1$}} & & \centering \textbf{2.2M (85.65\%)} &\centering \bm{$1.9\times 10^8$} \textbf{(38.71\%)} & \multirow{2}{1.0cm}{\centering \textbf{91.9\%}}& \multirow{2}{1.0cm}{\centering \textbf{-}} \\
          \cline{2-2}\cline{4-7}
          & \multirow{2}{2.2cm}{\centering \textbf{Sensitiveness-$\mathscr{C}_2$}} & & \centering \textbf{2.2M (85.65\%)} &\centering \bm{$1.9\times 10^8$} \textbf{(38.71\%)} & \multirow{2}{1.0cm}{\centering \textbf{91.9\%}}& \multirow{2}{1.0cm}{\centering \textbf{-}} \\
          \cline{2-2}\cline{4-7}
          & \multirow{2}{2.2cm}{\centering \textbf{Sensitiveness-$R$}} & & \centering \textbf{2.2M (85.65\%)} &\centering \bm{$1.9\times 10^8$} \textbf{(38.71\%)} & \multirow{2}{1.0cm}{\centering \textbf{92.0\%}}& \multirow{2}{1.0cm}{\centering \textbf{-}} \\
          \hline
    \end{tabular}}%
  \label{VGG16-results}%
\end{table*}%
     \renewcommand\arraystretch{1.0}
\begin{table*}[htb]
  \centering
  \scriptsize
  \caption{Results Of Different Pruning Methods For VGG-4 On MNIST.}
      \centering
      \setlength{\tabcolsep}{4.0mm}{
    \begin{tabular}{|p{2.5cm}|p{2.2cm}|p{1.0cm}|p{1.5cm}|p{1.5cm}|p{1.0cm}|p{1.0cm}|}
    \hline
    \multirow{2}{2.5cm}{\centering Models(datasets)} & \multirow{2}{2.2cm}{\centering Methods} & \centering Filters (Pruned) & \centering Parameters (Pruned) & \centering FLOPs (MACs) (Pruned)  &  \multirow{2}{1.0cm}{\centering Accuracy} \multirow{2}{1.0cm}&{\centering Constraints} \\
    \hline
    VGG-4(Baseline) &       &       & \centering 14.4M & \centering $3.2\times 10^8$ & \multirow{1}{1.0cm}{\centering 99.0\%}& \\
    \hline
    \multirow{12}[0]{*}{\centering VGG-4(MNIST)} &\multirow{2}{2.2cm}{\centering Filters Pruning} & \multirow{12}{1.0cm}{\centering 80\%}  & \centering 1.8M (87.50\%) & \centering $1.3\times10^8$ (59.38\%) &  \multirow{2}{1.0cm}{\centering 99.0\%}&\multirow{2}{1.0cm}{\centering -} \\
    \cline{2-2}\cline{4-7}
           & \multirow{2}{2.2cm}{\centering Taylor} &   & \centering 1.40M (90.28\%) & \centering $1.0\times 10^8$ (96.87\%) &  \multirow{2}{1.0cm}{\centering 98.5\%}&\multirow{2}{1.0cm}{\centering -} \\
          \cline{2-2}\cline{4-7}
          & \multirow{2}{2.2cm}{\centering Network Slimming} & &\centering 1.45M (89.93\%) &\centering $1.1\times 10^8$ (65.63\%) & \multirow{2}{1.0cm}{\centering 99.0\%}&\multirow{2}{1.0cm}{\centering Yes} \\
          \cline{2-2}\cline{4-7}
          & \multirow{2}{2.2cm}{\centering \textbf{Sensitiveness-$\mathscr{C}_1$}} & & \centering \textbf{1.37M (90.49\%)} &\centering \bm{$1.0\times 10^8$} \textbf{(96.87\%)} & \multirow{2}{1.0cm}{\centering \textbf{99.0\%}}&\multirow{2}{1.0cm}{\centering \textbf{-}} \\
          \cline{2-2}\cline{4-7}
          & \multirow{2}{2.2cm}{\centering \textbf{Sensitiveness-$\mathscr{C}_2$}} & & \centering \textbf{1.37M (90.49\%)} &\centering \bm{$1.0\times 10^8$} \textbf{(96.87\%)} & \multirow{2}{1.0cm}{\centering \textbf{99.0\%}}&\multirow{2}{1.0cm}{\centering \textbf{-}} \\
          \cline{2-2}\cline{4-7}
          & \multirow{2}{2.2cm}{\centering \textbf{Sensitiveness-$R$}} & & \centering \textbf{1.37M (90.49\%)} &\centering \bm{$1.0\times 10^8$} \textbf{(96.87\%)} & \multirow{2}{1.0cm}{\centering \textbf{99.0\%}}&\multirow{2}{1.0cm}{\centering \textbf{-}} \\
          \hline
    \end{tabular}}%
  \label{VGG4-results}%
\end{table*}%
 \renewcommand\arraystretch{1.0}
\begin{table*}[htb]
  \centering
  \scriptsize
  \caption{Results Of Different Pruning Methods For ResNet-18 On CIFAR-100.}
      \centering
      \setlength{\tabcolsep}{4.0mm}{
    \begin{tabular}{|p{2.5cm}|p{2.2cm}|p{1.0cm}|p{1.5cm}|p{1.5cm}|p{1.0cm}|p{1.0cm}|}
    \hline
    \multirow{2}{2.5cm}{\centering Models(datasets)} & \multirow{2}{2.2cm}{\centering Methods} & \centering Filters (Pruned) & \centering Parameters (Pruned) & \centering FLOPs (MACs) (Pruned)  &  \multirow{2}{1.0cm}{\centering Accuracy}&\multirow{2}{1.0cm}{\centering Constraints} \\
    \hline
    ResNet-18(Baseline) &       &       & \centering 11.68M & \centering $4\times 10^7$ & \multirow{1}{1.0cm}{\centering 85.1\%}&\\
    \hline
    \multirow{12}[0]{*}{\centering ResNet-18(CIFAR-100)} &\multirow{2}{2.2cm}{\centering Filters Pruning} & \multirow{12}{1.0cm}{\centering 37\%}  & \centering 2.2M (81.16\%) & \centering $2.0\times10^7$ (50\%) &  \multirow{2}{1.0cm}{\centering 85.0\%}&\multirow{2}{1.0cm}{\centering -}\\
    \cline{2-2}\cline{4-7}
           & \multirow{2}{2.2cm}{\centering Taylor} &   & \centering 2.0M (82.88\%) & \centering $2.0\times 10^7$ (50\%) &  \multirow{2}{1.0cm}{\centering 84.1\%}&\multirow{2}{1.0cm}{\centering -} \\
          \cline{2-2}\cline{4-7}
          & \multirow{2}{2.2cm}{\centering Network Slimming} & &\centering 2.1M (82.02\%) &\centering $2.0\times 10^7$ (50\%) & \multirow{2}{1.0cm}{\centering 85.0\%}&\multirow{2}{1.0cm}{\centering Yes} \\
          \cline{2-2}\cline{4-7}
          & \multirow{2}{2.2cm}{\centering \textbf{Sensitiveness-$\mathscr{C}_1$}} & & \centering \textbf{2.02M (82.79\%)} &\centering \bm{$2.0\times 10^7$} \textbf{(50\%)} & \multirow{2}{1.0cm}{\centering \textbf{85.0\%}}&\multirow{2}{1.0cm}{\centering \textbf{-}} \\
          \cline{2-2}\cline{4-7}
           & \multirow{2}{2.2cm}{\centering \textbf{Sensitiveness-$\mathscr{C}_2$}} & & \centering \textbf{2.02M (82.79\%)} &\centering \bm{$2.0\times 10^7$} \textbf{(50\%)} & \multirow{2}{1.0cm}{\centering \textbf{85.0\%}}&\multirow{2}{1.0cm}{\centering \textbf{-}} \\
           \cline{2-2}\cline{4-7}
            & \multirow{2}{2.2cm}{\centering \textbf{Sensitiveness-$R$}} & & \centering \textbf{2.02M (82.79\%)} &\centering \bm{$2.0\times 10^7$} \textbf{(50\%)} & \multirow{2}{1.0cm}{\centering \textbf{85.1\%}}&\multirow{2}{1.0cm}{\centering \textbf{-}} \\
          \hline
    \end{tabular}}%
  \label{ResNet18-results}%
\end{table*}%
  \subsection{Comparison of Different Pruning Criteria}
    In this section, we compared different pruning methods on VGG-16, Conv4 and ResNet-18 when they were fully trained with CIFAR-10, MNIST and CIFAR-100 correspondingly. The results were showed on Table \ref{VGG16-results}, \ref{VGG4-results} and \ref{ResNet18-results}. As described in section \ref{introduction}, "Filters pruning", "Taylor" method and "Network Slimming" were all weight value based structured pruning methods, which could only perform well when the original model provided efficient parameters. Notably, there was no obvious restriction for "Filters Pruning"\cite{haoli:pffec} and "Taylor Expansion" \cite{pavlo:nvidia} methods, which meant they were flexible to be applied across various common networks, such as pure chain-structure "VGGNets" and skip-structure "ResNets". However, "Network Slimming" \cite{zhuangliu:networkslimming} method was constrained by the output value of BatchNorm neurons, which could not evaluate each filter when there were no BatchNorm layers. In order to verify the effectiveness of our algorithm, the pruning ratio was set appropriately that could maintain the baseline accuracy. For VGG-16 on CIFAR-10, "Filters Pruning" method outperformed baseline result with 0.2\% rising of accuracy. However, the reduction of parameters was relative lower than others, and the pruning ratio of each layer was hard to determine. "Taylor Expansion" based method yielded impressive reduction both on parameters and FLOPs while with huge decrease in accuracy. "Network Slimming" showed excellent reduction both on parameters and FLOPs, as well as maintaining baseline accuracy. Our algorithm achieved comparable pruning results as "Network Slimming" while without the constraints of BatchNorm neurons, which could be implemented flexibly. Besides, our algorithm showed competitive pruning results with other methods on customized Conv-4 and ResNet-18. Both parameters and FLOPs could be reduced efficiently with sufficient flexibility.

 	\subsection{comparison of random and norm-based pruning strategies}
      In this section, we designed extra two experiments that were based on sensitiveness to remove filters by $\mathscr{C}_1$-norm and $\mathscr{C}_2$-norm, which could illustrate the advantage of random pruning that we introduced during the whole pruning process, as well as underlining the dominated status of network structure. At first, we would evaluate the sensitiveness of each layer according to the Equation (\ref{sensitiveness_calculations}) and Equation (\ref{average_sensitiveness}), then the filters will be removed from the original model with $\mathscr{C}_1$-norm, $\mathscr{C}_2$-norm, and random pruning, respectively. The pruning results were pictured as Fig. \ref{tab:comparision}. It was clear that there was no obvious difference for three different pruning strategies, and they all could reach up baseline accuracy after careful training. We renamed above three pruning strategies as sensitiveness-$\mathscr{C}_1$, sensitiveness-$\mathscr{C}_2$ and sensitiveness-$R$, and the compression details were illustrated on Table \ref{VGG16-results}, \ref{VGG4-results} and \ref{ResNet18-results}, respectively. Clearly, random strategy performed better than other two norm based strategies. Therefore, we could claim that the structure of the network dominated the representation ability.

\section{Analysis for pruning details}
  \subsection{Relationship to Automatic Pruning Strategy}
    Automatic pruning that were done by \cite{pavlo:nvidia} and \cite{zhuangliu:networkslimming} could remove filters globally through "Taylor Expansion" and "Network Slimming". In this part, we reviewed the pruning results for VGG-16 on CIFAR-10 and unveiled the structure of the pruned models when the pruning ratio was set to 65\%. For the "Taylor Expansion" method, pruning process would be implemented 5 times, which meant there were 549 filters being removed from the whole network with total 4224 filters during each iteration, then we retrained the pruned model with 10 epochs while revising the training epochs to 100 during last iteration. For "Network Slimming" method, we retrained the final compact model with 100 epochs to recover the accuracy. Fig. \ref{tab:taylor_sensitiveness}. showed the final retraining process. Apparently, sensitiveness based pruning and "Network Slimming" methods yielded better recovering accuracy than the method based on "Taylor expansion". The specific filters pruning for VGG-16 on CIFAR-10 were listed on Table \ref{tab:tablecompare}. It should be pointed out that the deeper layers were prone to be slimmed for three methods. Especially, "Network Slimming" and SBPF both maintained more filters among Conv2-Conv7, on the contrast, "Taylor Expansion" based method would abandon more filters in these layers. Table \ref{VGG16-results} compared the performance of above three pruning methods. When the pruning ratio was 65\%, our method and "Network Slimming" methods both could recover baseline accuracy, and "Network Slimming" performed well slightly while was constrained by BatchNorm neurons. On the contrast, "Taylor Expansion" could achieve better compression on parameters and FLOPs, but the pruned model almost lost 2\% accuracy, which only could be competed with our method when the original model was pruned with 80\% pruning ratio. The efficiency of our algorithm could be proved broadly.
    \subsection{Effect of Control Factor $\gamma$}
    The influence of $\gamma$ would be discussed on Supplementary material when VGG-16 was slimmed. For $\gamma=\frac{2}{3}$, all convolutional layers for VGG-16 could be classified into four groups, then we would discuss the layers grouping when $\gamma$ was set with different value. When $\gamma=\frac{1}{2}$, Conv2 still presented the highest score, and other layers showed similar scores as previous experiment. Therefore, the layers could be classified identically as $\gamma=\frac{2}{3}$. When we set $\gamma=\frac{3}{4}$, Conv2 was easy to be picked up as the most sensitive layer while the distinction among Conv1, Conv3, Conv4, Conv5, Conv6, Conv7, Conv8 was not obvious, which disturbed us to classify layers into appropriate groups. Finally, we set $\gamma=\frac{1}{4}$, which meant the stability would dominate the evaluation of sensitiveness. Notably, the scores across shallow layers were quite different, which assisted us to distinguish the sensitiveness of each layer. However, it was hard to group layers due to the large score gap across whole network. Besides, the difference among deep layers were small, and it could not reflect the real importance within these layers. Therefore, it was reasonable to authorize reliability more weights than stability.
   \subsection{Exploration of efficiency for our method}
    The efficiency of our method could be supported by the explanation that filters in insensitive layers were less important and would impose side effect for representation ability of original model, as well as bringing about overfitting problems. Therefore, slimming appropriately for these insensitive layers could not result in accuracy hurting while contributing to improve the power of representation for original model. On the contrary, sensitive layers could maintain more important structured information, and tiny change of structure may cause accuracy hurt dramatically. At the same time, we explored the FLOPs occupation of each convolutional layer for VGG-16 on CIFAR-10 in Supplementary material. It was clear that there was no obvious relationship between FLOPs and sensitiveness, then more filters within the layers that occupied the most FLOPs should be removed when the sensitiveness was similar. Additionally, the retraining process could update gradient information and carry out a group of new parameters, which strengthen the view that representation ability of networks was dominated by structure rather weight information.

However, there are some apparent difficulty for applying our algorithm on network pruning tasks. Especially, the confirmation of pruning ratio for each layer is still determined by hand, which is cumbersome to find the optimal compact model. Therefore, some automatic pruning strategies could be used to find a better pruned model.

\section{Conclusion}
We proposed a sensitiveness based pruning framework with multi-steps to slim VGG-16, Conv-4 and ResNet-18 across various datasets. The experiments showed that the evaluation of sensitiveness could maintain consistent even when the original model was not trained well, which meant our pruning criteria was robust to the update of parameters for individual filter. Meanwhile, we strengthen the view that the structure of network dominated representation ability, which motivate us to focus on reducing the size of network at layer-level. However, one of the main difficulty in our algorithm was the confirmation of pruning ratio in the final iterative pruning process. It was hard to determine exact pruning ratio for each layer through once time. Therefore, we will utilize automatic techniques to be combined with sensitiveness criteria to look for better pruned architecture in the future.

\footnotesize
\bibliographystyle{IEEEtran}
\bibliography{IEEEexample}

\begin{thebibliography}{10}
\providecommand{\url}[1]{#1}
\csname url@samestyle\endcsname
\providecommand{\newblock}{\relax}
\providecommand{\bibinfo}[2]{#2}
\providecommand{\BIBentrySTDinterwordspacing}{\spaceskip=0pt\relax}
\providecommand{\BIBentryALTinterwordstretchfactor}{4}
\providecommand{\BIBentryALTinterwordspacing}{\spaceskip=\fontdimen2\font plus
\BIBentryALTinterwordstretchfactor\fontdimen3\font minus
  \fontdimen4\font\relax}
\providecommand{\BIBforeignlanguage}[2]{{%
\expandafter\ifx\csname l@#1\endcsname\relax
\typeout{** WARNING: IEEEtran.bst: No hyphenation pattern has been}%
\typeout{** loaded for the language `#1'. Using the pattern for}%
\typeout{** the default language instead.}%
\else
\language=\csname l@#1\endcsname
\fi
#2}}
\providecommand{\BIBdecl}{\relax}
\BIBdecl

\bibitem{kriz:ima}
A.~Krizhevsky, I.~Sutskever, and G.~Hinton, ``Imagenet classification with deep
  convolutional neural networks,'' in \emph{Proc. Adv. Neural Inf. Process.
  Syst.}, Lake Tahoe, Dec. 2012, pp. 1106--1114.

\bibitem{googlenet}
C.~Szegedy, W.~Liu, Y.~Jia1, P.~Sermanet1, and S.~Reed, ``Going deeper with
  convolutions,'' in \emph{Proc. IEEE/CVF Conf. Comput. Vis. Pattern
  Recognit.}, Boston, Massachusetts, Jun. 2015, pp. 1--9.

\bibitem{googlenetv4}
C.~Szegedy, C.~Ioffe, S.~Vanhoucke, V.~Alemi, and A.~Alexander, ``Inception-v4,
  {I}nception-{R}es{N}et and the impact of residual connections on learning,''
  in \emph{Proc. Assoc. Adv. Art. Intel.}, Phoenix, Arizona, Feb. 2016, pp.
  4278--4284.

\bibitem{densnet}
G.~Huang, Z.~Liu, L.~Maaten, and K.~Weinberger, ``Densely connected
  convolutional networks,'' in \emph{Proc. IEEE/CVF Conf. Comput. Vis. Pattern
  Recognit.}, Hawaii, Feb. 2017, pp. 2261--2269.

\bibitem{girshick-et-al:richfeature}
R.~Girshick, J.~Donahue, T.~Darrell, and J.~Malik, ``Rich feature hierarchies
  for accurate object detection and semantic segmentation,'' in \emph{Proc.
  IEEE/CVF Conf. Comput. Vis. Pattern Recognit.}, Columbus,Ohio, Jun. 2014, pp.
  580--587.

\bibitem{fasterrcnn}
S.~Ren, K.~He, R.~Girshic, and J.~Sun, ``Faster {R-CNN}: Towards real-time
  object detection with region proposal networks,'' \emph{IEEE Trans. Pattern
  Anal. Mach. Intell.}, vol.~39, no.~6, pp. 1137--1149, Dec. 2017.

\bibitem{maskrcnn}
K.~He, G.~Gkioxari, P.~Dolar, and R.~Girshick, ``Mask {R-CNN},'' in \emph{IEEE
  Int. Conf. Comput. Vis. (ICCV)}, Venice, Italy, Oct. 2017, pp. 2980--2988.

\bibitem{jonathan-et-al:fcnss}
J.~Long, E.~Shelhamer, and T.~Darrell, ``Fully convolutional networks for
  semantic segmentation,'' in \emph{Proc. IEEE/CVF Conf. Comput. Vis. Pattern
  Recognit.}, Boston, Massachusetts, Jun. 2015, pp. 3431--3440.

\bibitem{hekaiming:resnet}
K.~He, X.~Zhang, S.~Ren, and J.~Sun, ``Deep residual learning for image
  recognition,'' in \emph{Proc. IEEE/CVF Conf. Comput. Vis. Pattern Recognit.},
  Las Vegas, Jun. 2016, pp. 770--778.

\bibitem{karen:vdc}
K.~Simonyan and A.~Zissermann, ``Very deep convolutional networks for
  large-scale image recognition,'' in \emph{Proc. Int. Conf. Learn.
  Represent.}, San Diego, May 2015.

\bibitem{Yann:obd}
Y.~LeCun, J.~Denker, and S.~Solla, ``Optimal brain damage,'' in \emph{Proc.
  Int. Conf. Mach. Learn.}, Denver, Colorado, Nov. 1990, pp. 598--605.

\bibitem{babak:obu}
B.~Hassibi and D.~Stork, ``Second order derivatives for network pruning:
  Optimal brain surgeon,'' in \emph{Proc. Adv. Neural Inf. Process. Syst.},
  Denver, Colorado, Nov. 1993, pp. 164--171.

\bibitem{sparsity:soravit}
S.~Changpinyo, M.~Sandler, and A.~Zhmoginov, ``The power of sparsity in
  convolutional neural networks,'' in \emph{arXiv:1702.06257}, 2017.

\bibitem{nn-filter-grouping}
Q.~Guo, X.~Wu, J.~Kittler, and Z.~Feng, ``Self-grouping convolutional neural
  networks,'' \emph{Neural. Netw.}, vol. 132, pp. 491--505, Dec. 2020.

\bibitem{haoli:pffec}
H.~Li, A.~Kadav, I.~Durdanovic, H.~Samet, and H.~Graf, ``Pruning filters for
  efficient convnets,'' in \emph{Proc. Int. Conf. Learn. Represent.}, Toulon,
  France, Apr. 2017.

\bibitem{hengyuan:datadriven}
H.~Hu, R.~Peng, Y.~Tai, and C.~Tang, ``Network trimming: A data-driven neuron
  pruning approach towards efficient deep architectures,'' in
  \emph{arXiv:1607.03250}, Jul. 2016.

\bibitem{pavlo:nvidia}
P.~Molchanov, S.~Tyree, T.~Karras, T.~Aila, and J.~Kautz, ``Pruning
  convolutional neural networks for resource efficient inference,'' in
  \emph{arXiv:1611.06440}, 2016.

\bibitem{zhuangliu:networkslimming}
Z.~Liu, J.~Li, Z.~Shen, G.~Huang, S.~Yan, and C.~Zhang, ``Learning efficient
  convolutional networks through network slimming,'' in \emph{IEEE Int. Conf.
  Comput. Vis. (ICCV)}, Venice, Italy, Oct. 2017, pp. 2755--2763.

\bibitem{zehao:datadriven}
Z.~Huang and N.~Wang, ``Data-driven sparse structure selection for deep neural
  networks,'' in \emph{Proc. Eur. Conf. Comput. Vis.}, Munich, Germany, Sep.
  2018, pp. 317--334.

\bibitem{visualizing}
M.~Li, S.~Wang, and Q.~Zhang, ``Visualizing the emergence of intermediate
  visual patterns in dnns,'' in \emph{Proc. Adv. Neural Inf. Process. Syst.},
  Dec. 2021, pp. 6594--6607.

\bibitem{datafree:suraj}
S.~Srinivas and R.~Babu, ``Data-free parameter pruning for deep neural
  networks,'' in \emph{Proc. Bri. Mach. Vis. Conf.}, Swansea, UK, Sep. 2015,
  pp. 31.1--31.12.

\bibitem{song:cdnnptqhc}
S.~Han, H.~Mao, and W.~Dally, ``Deep compression: Compressing deep neural
  networks with pruning, trained quantization and huffman coding,'' in
  \emph{Proc. Int. Conf. Learn. Represent.}, San Juan, Puerto Rico, May 2016.

\bibitem{lsnn}
C.~Louizos, M.~Welling, and D.~Kingma, ``Learning sparse neural networks
  through $l_0$ regularization,'' in \emph{Proc. Int. Conf. Learn. Represent.},
  Vancouver, Canada, Oct. 2018.

\bibitem{dropoutsaprse}
D.~Molchanov, A.~Ashukha, and D.~Vetrov, ``Variational dropout sparsifies deep
  neural networks,'' in \emph{Proc. Int. Conf. Mach. Learn.}, Sydney,
  Australia, Aug. 2017.

\bibitem{decomposition}
K.~Wu, Y.~Guo, and C.~Zhang, ``Compressing deep neural networks with sparse
  matrix factorization,'' \emph{IEEE Trans. Neural Netw. Learn Syst.}, vol.~31,
  no.~10, pp. 3828--3838, Oct. 2020.

\bibitem{lottery}
F.~Jonathan and C.~Michael, ``The lottery ticket hypothesis: Finding sparse,
  trainable neural networks,'' in \emph{Proc. Int. Conf. Learn. Represent.},
  New Orleans, Louisiana, May 2019.

\bibitem{song:eie}
S.~Han, X.~Liu, H.~Mao, J.~Pu, A.~Pedram, M.~Horowitz, and W.~Dally, ``{EIE}:
  Efficient inference engine on compressed deep neural network,'' in
  \emph{Proc. Int. Symp. Comput. Archit.}, Seoul, Korea, Jun. 2016, pp.
  243--254.

\bibitem{NISP}
R.~Yu, A.~Li, C.~Chen, J.~Lai, V.~Morariu, and X.~Han, ``Nisp: Pruning networks
  using neuron importance score propagation,'' in \emph{Proc. IEEE/CVF Conf.
  Comput. Vis. Pattern Recognit.}, Salt Lake City, June 2018, pp. 9194--9203.

\bibitem{Thinet}
J.~Luo, J.~Wu, and W.~Lin, ``Thinet: A filter level pruning method for deep
  neural network compression,'' in \emph{IEEE Int. Conf. Comput. Vis. (ICCV)},
  Venice, Italy, Oct. 2017, pp. 5068--5076.

\bibitem{hanzhou:tcn}
H.~Zhou, J.~Alvarez, and F.~Porikli, ``Less is more: Towards compact cnns,'' in
  \emph{Proc. Eur. Conf. Comput. Vis.}, Amsterdan, Netherland, Oct. 2016, pp.
  662--677.

\bibitem{nn-decom}
B.~Wu, D.~Wang, G.~Zhao, L.~Deng, and G.~Li, ``Hybrid tensor decomposition in
  neural network compression,'' \emph{Neural. Netw.}, vol. 132, pp. 209--320,
  Dec. 2020.

\bibitem{nn-structured}
L.~Wen, X.~Zhang, H.~Bai, and Z.~Xu, ``Structured pruning of recurrent neural
  networks through neuron selection,'' \emph{Neural. Netw.}, vol. 123, pp.
  134--141, Mar. 2020.

\bibitem{metalearning}
Z.~Liu, H.~Mu, X.~Zhang, Z.~Guo, X.~Yang, K.~Cheng, and J.~Sun,
  ``Meta{P}runing: Meta learning for automatic neural network channel
  pruning,'' in \emph{IEEE Int. Conf. Comput. Vis. (ICCV)}, Seoul, Korea, Oct.
  2019, pp. 3295--3304.

\bibitem{NullHop}
A.~Aimar, H.~Mostafa, E.~Calabrese, A.~Rios-Navarro, R.~Tapiador-Morales,
  I.~Lungu, M.~B. Milde, F.~Corradi, A.~Linares-Barranco, S.~Liu, and
  T.~Delbruck, ``Null{H}op: A flexible convolutional neural network accelerator
  based on sparse representations of feature maps,'' \emph{IEEE Trans. Neural
  Netw. Learn Syst.}, vol.~30, no.~3, pp. 644--656, Mar. 2019.

\bibitem{coprocessor}
N.~Shah, P.~Chaudhari, and K.~Varghese, ``Runtime programmable and memory
  bandwidth optimized fpga-based coprocessor for deep convolutional neural
  network,'' \emph{IEEE Trans. Neural Netw. Learn Syst.}, vol.~29, no.~12, pp.
  5922--5934, Dec. 2018.

\bibitem{linear:denton}
E.~Denton, W.~Zaremba, J.~Bruna, Y.~LeCun, and R.~Fergus, ``Exploiting linear
  structure within convolutional networks for efficient evaluation,'' in
  \emph{Proc. Adv. Neural Inf. Process. Syst.}, Montreal, QC, Dec. 2014.

\bibitem{approximation}
R.~Cintra, S.~Duffner, C.~Garcia, and A.~Leite, ``Low-complexity approximate
  convolutional neural networks,'' \emph{IEEE Trans. Neural Netw. Learn Syst.},
  vol.~29, no.~12, pp. 5981--5992, Dec. 2018.

\bibitem{quantization:weilin}
W.~Chen, J.~Wilson, S.~Tyree, K.~Weinberger, and Y.~Chen, ``Compressing neural
  networks with the hashing trick,'' in \emph{Proc. Int. Conf. Mach. Learn.},
  Lille, France, Jul. 2015, pp. 2285--2294.

\bibitem{binary:matthiew}
M.~Courbariaux, I.~Hubara, D.~Soudry, R.~Yaniv, and Y.~Bengio, ``Binary{N}et:
  Training deep neural networks with weights and activations constrained to +1
  or -1,'' in \emph{arXiv:1602.02830}, 2016.

\bibitem{knowledgedistation}
G.~Hinton, O.~Vinyals, and J.~Dean, ``Distilling the knowledge in a neural
  network,'' in \emph{arXiv:1503.02531}, 2015.

\bibitem{knowledgedistationbaysion}
A.~Korattikara, V.~Rathod, K.~Murphy, and M.~Welling, ``Bayesian dark
  knowledge,'' in \emph{Proc. Adv. Neural Inf. Process. Syst.}, Montreal,
  Canada, Dec. 2015, pp. 3438--3446.

\bibitem{MobileNetV1}
A.~Howard, M.~Zhu, B.~Chen, D.~Kalenichenko, and W.~Wang, ``Mobile{N}ets:
  Efficient convolutional neural networks for mobile vision applications,'' in
  \emph{arXiv:1704.04861}, 2017.

\bibitem{MobileNetV2}
M.~Sandler, A.~Howard, M.~Zhu, A.~Zhmoginov, and L.~Chen, ``Mobile{N}et{V}2:
  Inverted residuals and linear bottlenecks,'' in \emph{Proc. IEEE/CVF Conf.
  Comput. Vis. Pattern Recognit.}, Salt Lake City, June 2018, pp. 4510--4520.

\bibitem{MobileNetV3}
A.~Howard, M.~Sandler, G.~Chu, L.~Chen, B.~Chen, and M.~Tan, ``Searching for
  {M}obile{N}et{V}3,'' in \emph{Proc. IEEE/CVF Conf. Comput. Vis. Pattern
  Recognit.}, Long Beach, June 2019, pp. 1314--1324.

\bibitem{NASreinforce}
B.~Zoph and Q.~Le, ``Neural architecture search with reinforcement learning,''
  in \emph{Proc. Int. Conf. Learn. Represent.}, Vancouver, Canada, Oct. 2018.

\bibitem{NASlearningtransferingable}
B.~Zoph, V.~Vasudevan, J.~Shlens, and Q.~Le, ``Learning transferable
  architectures for scalable image recognition,'' in \emph{Proc. IEEE/CVF Conf.
  Comput. Vis. Pattern Recognit.}, Salt Lake City, Jun. 2018, pp. 8697--8710.

\bibitem{vFM}
M.~Hasnat, J.~Bohne, S.~G. J.~Milgram, and L.~Chen, ``Von mises-fisher mixture
  model-based deep learning: Application to face verification.'' in \emph{arXiv
  preprint arXiv:1706.04264}, 2017.

\end{thebibliography}

\end{document}